\documentclass[lettersize,journal]{IEEEtran}
\usepackage{amsmath,amsfonts}
\usepackage{algorithm}
\usepackage{array}
\usepackage[caption=false,font=normalsize,labelfont=sf,textfont=sf]{subfig}
\usepackage{textcomp}
\usepackage{stfloats}
\usepackage{url}
\usepackage{verbatim}
\usepackage{graphicx}
\usepackage{subcaption}
\usepackage{cite}
\usepackage{multirow}
\usepackage{hyperref}
\usepackage{algpseudocode}
\usepackage{xcolor}
\usepackage{multirow}
\usepackage{tabularx} 
\usepackage{caption} 
\usepackage{booktabs} 
\usepackage{soul}
\usepackage{adjustbox}

\usepackage[normalem]{ulem} 

\usepackage{float}  

\usepackage{bbding}
\usepackage{pifont}
\begin{document}

\title{Rectifying Adversarial Sample with Low Entropy Prior for Test-Time Defense}

\author{Lina Ma, Xiaowei Fu, Fuxiang Huang, Xinbo Gao,~\IEEEmembership{Fellow,~IEEE}, and Lei Zhang,~\IEEEmembership{Senior Member,~IEEE}
\thanks{This work was partially supported by National Natural Science Fund of China (62271090), National Key R\&D Program of China (2021YFB3100800),  Chongqing Natural Science Fund, and National Youth Talent Project. \textit{(L. Ma and X. Fu contribute equally) (Corresponding author: Lei Zhang)}
}
\thanks{L. Ma, X. Fu  and L. Zhang are with Chongqing Key Laboratory of Bio-perception and Multimodal Intelligent Information Processing, Chongqing University and the School of Microelectronics and Communication Engineering, Chongqing University, Chongqing 400044, China. (E-mail: mln@stu.cqu.edu.cn, xwfu@cqu.edu.cn, leizhang@cqu.edu.cn)}
\thanks{F. Huang is with Chongqing Key Laboratory of Bio-perception and Multimodal Intelligent Information Processing, Chongqing University, Chongqing 400044, China, and Department of Computer Science and Engineering, The Hong Kong University of Science and Technology, Hong Kong. (E-mail: huangfuxiang@cqu.edu.cn)}
\thanks{X. Gao is with the Chongqing Key Laboratory of Image Cognition, Chongqing University of Posts and Telecommunications, Chongqing 400065, China. (E-mail: gaoxb@cqupt.edu.cn)}
\thanks{Manuscript received April 19, 2005; revised August 26, 2015.}}

\markboth{Journal of \LaTeX\ Class Files,~Vol.~14, No.~8, August~2021}%
{Shell \MakeLowercase{\textit{et al.}}: A Sample Article Using IEEEtran.cls for IEEE Journals}


\maketitle

\begin{abstract}
Existing defense methods fail to defend against unknown attacks and thus raise generalization issue of adversarial robustness. 
To remedy this problem, we attempt to delve into some underlying common characteristics among various attacks for generality. In this work, we reveal the commonly overlooked low entropy prior (LE) implied in various adversarial samples, and shed light on the universal robustness against unseen attacks in inference phase. LE prior is elaborated as two properties across various attacks as shown in Fig. \ref{fig1} and \ref{fig2}: 1) low entropy misclassification for adversarial samples and 2) lower entropy prediction for higher attack intensity. This phenomenon stands in stark contrast to the naturally distributed samples. The LE prior can instruct existing test-time defense methods, thus we propose a two-stage REAL approach: \underline{Re}ctify \underline{A}dversarial sample based on \underline{L}E prior for test-time adversarial rectification. Specifically, to align adversarial samples more closely with clean samples, we propose to first rectify adversarial samples misclassified with low entropy by \textit{reverse} maximizing prediction entropy, thereby eliminating their adversarial nature. To ensure the rectified samples can be correctly classified with low entropy, we carry out secondary rectification by \textit{forward} minimizing prediction entropy, thus creating a Max-Min entropy optimization scheme. Further, based on the second property, we propose an attack-aware weighting mechanism to adaptively adjust the strengths of Max-Min entropy objectives. Experiments on several datasets show that REAL can greatly improve the performance of existing sample rectification models.

\end{abstract}


\section{Introduction}
\IEEEPARstart{A}{dversarial} attacks expose vulnerabilities of deep neural networks and arise the thinking about security issues of real application, including image classification \cite{goodfellow2014explaining,szegedy2013intriguing}, autonomous driving \cite{modas2020toward}, facial recognition \cite{zheng2023robust}, image captioning \cite{xu2019exact} and other fields \cite{xu2020adversarial}. To mitigate the dangers posed by adversarial attacks, adversarial defenses have emerged \cite{kurakin2022adversarial,madry2017towards,du2018enhancing,wang2019improving,wongfast,wang2023better}. However, most existing defense methods only perform well on known attacks but struggle to generalize unseen attacks, highlighting the generalization challenge in adversarial robustness \cite{rice2020overfitting,chen2020robust,tang2024robust,zhang2024meta}. As shown in our pilot experiment presented in Table \ref{table1}, adversarial training (AT), i.e., PGD-AT has a good defense effect against PGD (known attacks), but it is difficult to achieve effective defense against CW and DF (unseen attacks).

\begin{figure}[t]
\centering
\includegraphics[width=\columnwidth, height=0.5\columnwidth]{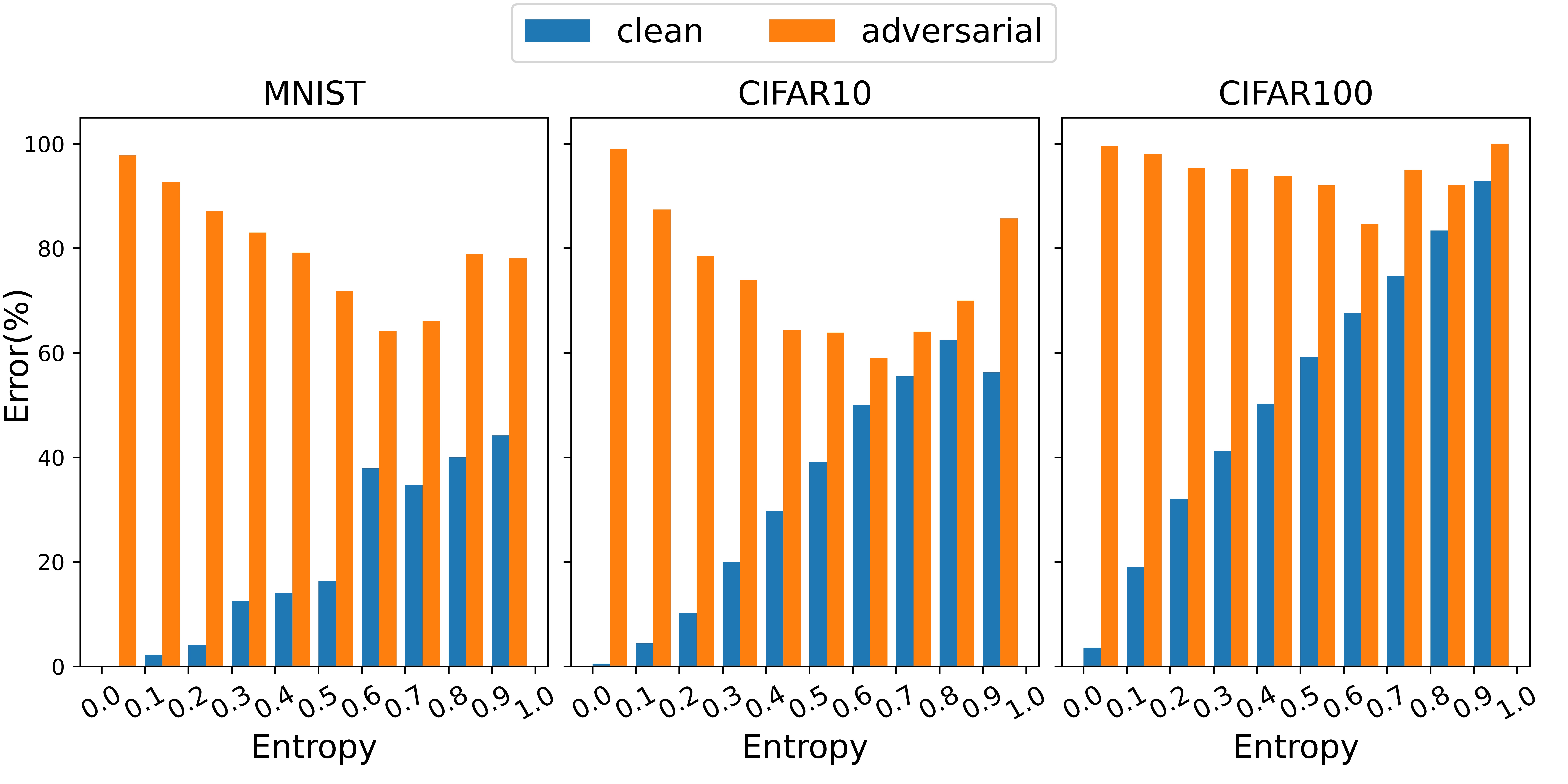}
\caption{The relationship between predicted entropy and error rates on adversarial samples generated by PGD attack and clean samples of three datasets. 
}
\label{fig1}
\end{figure}

\textcolor{black}{Recent studies \cite{wangtent,zhang2022memo} that focused on generalization in \textbf{\textit{natural robustness}} have inspired us to delve into generalization of adversarial robustness.} They follow a consensus that, for natural distribution samples, entropy is related to error rate. That is, for a well-trained model, statistical analysis on the predicted information entropy for samples reveals that lower entropy often corresponds to higher prediction confidence and lower error rates. As shown in Fig. \ref{fig1}. The blue bar chart represents the relationship between the predicted entropy of clean samples and their error rates. As entropy gradually increases, the error rate rises as well, exhibiting a proportional relationship between entropy and error rate. Building on the shared properties of entropy in natural samples, these works update model by minimizing the predicted entropy during testing to enhance the generalization in natural robustness. This inspiration leads us to think deeply:

\emph{Can we also leverage the shared entropy property of adversarial samples during testing for test-time defense thus achieving the generalization in \textbf{\textit{adversarial robustness}}?}

To answer the above question, we delve into the entropy of adversarial samples and reveal a prevalent yet frequently disregarded low-entropy prior (LE), consisting of two properties. \textbf{Property 1}: Low-entropy misclassification, i.e., the adversarial samples are frequently misclassified with high confidence, indicating a low entropy phenomenon. \textbf{Property 2}: Lower-entropy prediction for higher attack intensity, i.e., the entropy of the sample diminishes with escalating attack intensity.

\begin{figure}[t]
\centering
\includegraphics[width=\linewidth]{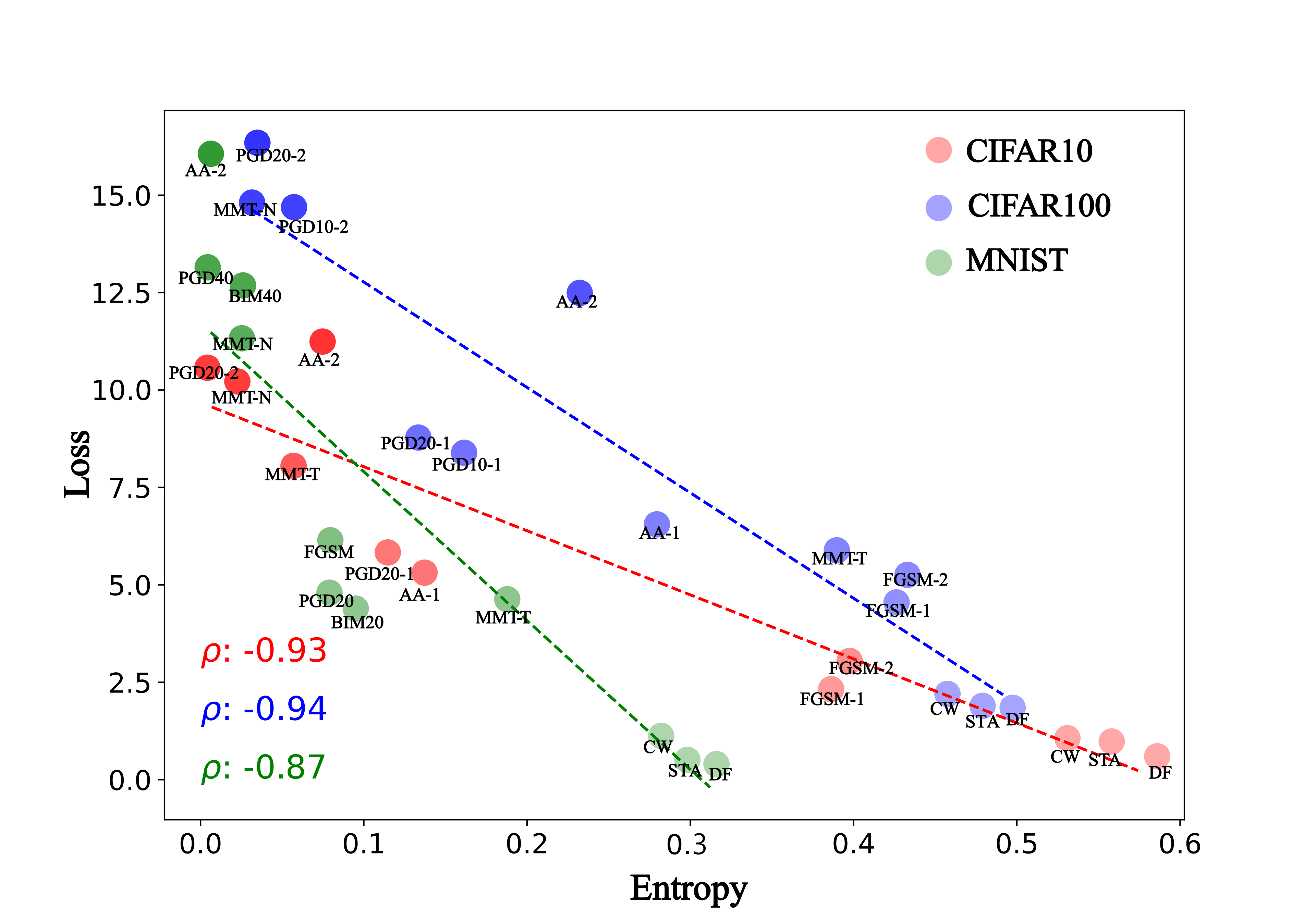} 
\caption{Greater attack strength causes larger loss and lower entropy across various attack methods (represented by different points) on three datasets. The -1 and -2 identifiers after the attack name represent different attack amplitudes, respectively 4/255, 8/255. $\rho$ means correlation coefficient.}
\label{fig2}
\end{figure}

\textcolor{black}{To interpret the LE prior, we firstly present the statistical relationship between entropy and error rate (defined as the proportion of misclassified samples in the model's prediction results) of adversarial samples generated using PGD attack \cite{madry2017towards} on MNIST, CIFAR10 and CIFAR100\cite{krizhevsky2009learning} in Fig. \ref{fig1}.} The yellow bars in the chart represent adversarial samples, where it can be observed that even with very low predicted entropy, the classification error rate remains high. This contrasts significantly with the proportional relationship between entropy and error rate observed in natural samples. Adversarial samples exhibit strong adversarial nature, i.e., they tend to be confidently misclassified with low prediction entropy. This corresponds to \textbf{Property 1}. 

Additionally, to achieve robust generalization against various attacks, we further present entropy statistics results for multiple types of attacks, aiming to explore the common properties of their entropy characteristics.
Specifically, we present more attack methods, such as CW, DF, FGSM, PGD, STA, MIM and AA \cite{carlini2017towards,moosavi2016deepfool,goodfellow2014explaining,madry2017towards,xiao2018spatially,dong2018boosting,croce2020reliable}. \textcolor{black}{As depicted in Fig. \ref{fig2}, we illustrate the relationship between the predicted entropy of adversarial samples under various attacks and their attack intensity (represented by classification loss, and a higher loss corresponds to a stronger attack)}. From Fig. \ref{fig2}, we observe variations in the entropy properties among adversarial samples generated by different attack methods. Most points in the figure are concentrated on the left side, indicating that most adversarial samples generated by various attacks exhibit low entropy. This observation aligns with the Property 1, which suggests a general tendency towards low entropy in adversarial samples. Certainly, there are instances where certain data points deviate to the lower right, which indicates that the predicted entropy value of adversarial samples is affected by the strength of attacker. We conduct an analysis of the correlation between the predicted entropy of adversarial samples and their attack intensity across three datasets. The relationship coefficient $\rho$ between predicted entropy and attack strength is calculated on each dataset, as shown in Fig. \ref{fig2}, revealing a significant negative correlation. The analysis results show that adversarial samples with higher classification loss (i.e., stronger attack intensity) often exhibit lower entropy. This pattern starkly contrasts with the natural distribution of samples and corresponds to \textbf{Property 2}. \textcolor{black}{Moreover, as shown in Fig. \ref{fig1}, the classification error rate of high-entropy samples (entropy $>$ 0.6) exhibits an upward trend, which aligns with the explanation provided by \textbf{Property 2}. This is because such samples have lower attack intensity, making them similar to natural samples, where entropy and error rate are positively correlated, thus validating the conclusion of \textbf{Property 2}.}

To defend against adversarial samples, aligning them with clean samples is crucial. Based on the \textbf{Property 1} of LE prior, that indicates the difference in entropy properties between adversarial and clean samples, a natural question arises: \textit{how to prevent adversarial samples from being misclassified with high confidence?} Regarding \textbf{Property 1} for adversarial samples, it instructs us to have a natural idea: transform or rectify these adversarial samples into mask sample by reverse maximizing their entropy, from LE to HE (high entropy). This \textbf{reverse rectification} step eliminates the strong adversarial nature of adversarial samples. But not just that, our ultimate goal is to obtain purified samples that can be classified correctly with high confidence as natural samples. Hence, we propose to further rectify the mask samples by minimizing entropy. This \textbf{forward rectification} step endows the rectified samples with benign nature. These two steps exploit the first property of LE prior and forms a Max-Min entropy optimization scheme for test-time sample rectification. Notably, this LE prior guided Max-Min entropy optimization mechanism is orthogonal to existing models, and thus can be easily plugged-played.  

Regarding \textbf{Property 2} as shown in Fig. \ref{fig2}, the attack intensity increases (i.e., increased loss) and the predicted entropy decreases. However, the above Max-Min entropy mechanism neglects to take into account the differences in strength among various attackers. 
So we propose an attack-aware weighted Max-Min entropy optimization mechanism, which exploits the attack intensity to dynamically weight the Max-Min entropy loss by assessing samples' predicted entropy. 
In nature, it dynamically assigns smaller weight of the Max-Min entropy loss for adversarial samples with lower attack intensity (such as those generated by DF attack), and vice versa. Furthermore, due to the operation of Max-Min entropy mechanism is similar to introducing a self-adversarial process, it increases the difficulty of sample rectification. Therefore, we propose a multi-step optimization algorithm that automatically regulates the number of purification steps during the purification process, thereby further enhancing the robustness performance of rectified samples.

Our contributions are summarized as follows.
\begin{itemize}
\item  We first find and harness the low entropy prior (LE) of adversarial samples to bolster the generalization capability of current defense methods against unseen attacks. 
\item We propose a simple yet effective two-stage \textit{reverse-forward} rectification approach: \underline{Re}ctified \underline{A}dversarial Sample based on \underline{L}oE prior (REAL) for Test-Time Defense,  with an attack-aware Max-Min entropy optimizer. 
\item The LE prior guided mechanism can instruct existing methods to improve universal adversarial robustness, and sufficient experiments verified the superiority.
\end{itemize}
\section{Related work}
\subsection{Information entropy}
Information entropy measures event uncertainty in information theory. Small prediction entropy indicates higher prediction certainty \cite{massey1994guessing}, and it is often associated with lower prediction error rate. \cite{wangtent,zhang2022memo} explore this relationship on natural samples and use this conclusion to achieve generalization in natural robustness during test-time adaptation through the minimization of information entropy. \cite{wang2021fighting} expands the application in the field of adversarial defense. In this research, we address the oversight in prior studies on prediction entropy by incorporating the low entropy prior (LE) of adversarial samples and propose a method for its effective utilization.
\subsection{Adversarial attack}
Since \cite{szegedy2013intriguing} discovers the vulnerability of neural networks: adding some disturbances to images that cannot be distinguished by the human eye can lead to images being misclassified with high confidence by the network. And some works have also emerged to explore the vulnerability of neural networks \cite{athalye2018synthesizing,xiao2018spatially,9599534,10065458,9430730}. 
The earliest adversarial attack, Fast Gradient Sign Method (FGSM), used single step gradients to create adversarial samples. Subsequently, in order to improve the effectiveness of adversarial samples, multi-step gradient update attack methods such as Projected Gradient Descent (PGD) \cite{madry2017towards}, Basic Iterative Methods (BIM) \cite{kurakin2022adversarial}, Momentum Iterative Method (MIM) \cite{dong2018boosting}, etc. were developed. Besides the above-mentioned gradient-based iterative update methods, there are also methods based on classification layer perturbation like DeepFool (DF) \cite{moosavi2016deepfool} and those based on conditional optimization such as the Carlini and Wagner attack (CW) \cite{carlini2017towards}. Additionally, Auto Attack (AA) \cite{croce2020reliable} frameworks that integrate two new versions of PGD, namely APGD-CE and APGD-DLR, in conjunction with Fast Adaptive Boosting Attack (FAB) \cite{croce2020minimally} and Square Attack \cite{andriushchenko2020square}, have been developed to assess adversarial robustness, collectively referred to as AutoAttack (AA). 
Adversarial samples generated by different attack methods perform differently and this raises the challenge of generalization in adversarial defense. 
\begin{figure*}[t]
\centering
\includegraphics[width=\linewidth]{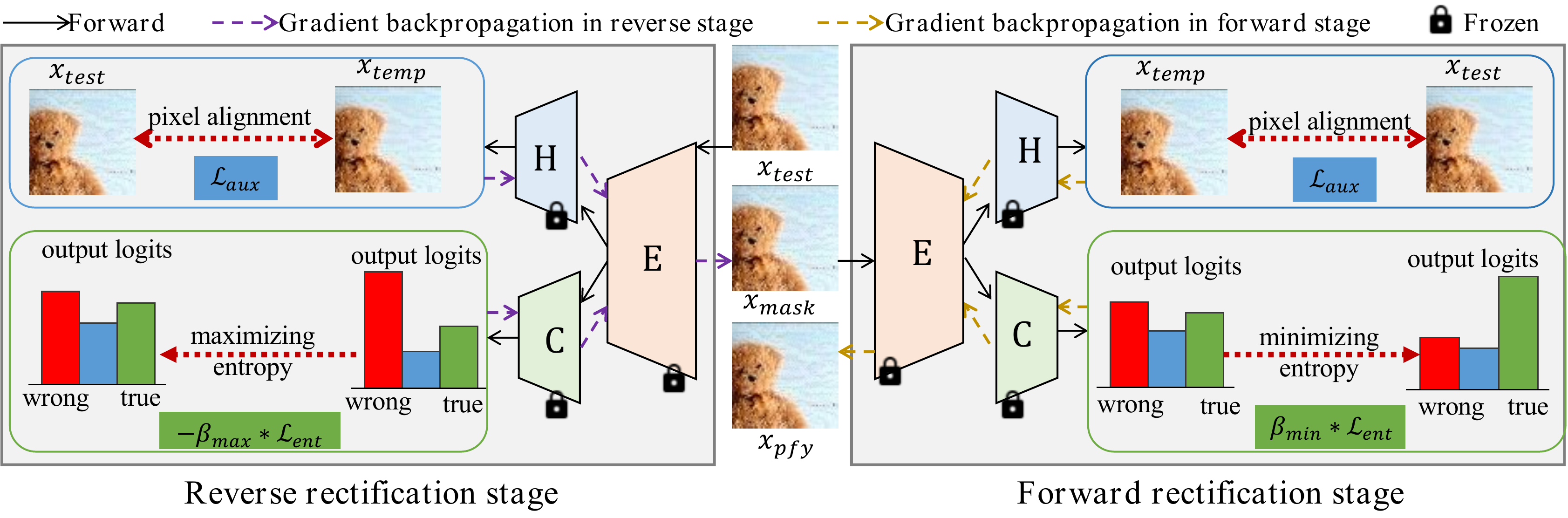}
\caption{Overall framework of REAL. During testing, images are rectified by optimizing the rectification loss. In stage 1, adversarial samples are disrupted by maximizing prediction entropy, resulting in masking effects (i.e. generating mask samples). In stage 2, enlightenment of mask samples is achieved through entropy minimization, resulting in final purified samples.}
\label{fig3}
\end{figure*}
\subsection{Adversarial defense}
Adversarial training (AT) \cite{kurakin2022adversarial,wongfast,dong2023enemy,amini2020towards,10542377} belong to be one of the most effective defense strategies. However, their generalization to unknown attackers is limited and is computationally expensive for large-scale networks. To effectively address these two challenges, test-time defense strategy is proposed, drawing inspiration from domain adaptation (DA) \cite{boudiaf2022parameter,wangtent,he2020momentum,sun2019test}. Adversarial rectification, aimed at purifying adversarial samples during inference, demonstrates robust generalization across various attack scenarios encountered during the testing phase \cite{lin2020dual}. Among them, some works purify samples by adding independent additional networks, such as using GAN-based models \cite{samangouei2018defense,9349536}, autoencoders \cite{hwang2019puvae,willettsimproving,yin2022defending,yang2021class}, transformer\cite{li2023learning} and the recently popular diffusion model \cite{nie2022diffusion,xiao2022densepure,chenrobust}. However, training these additional generative networks is time-consuming and laborious. Another part of these works combine self-supervised learning with lighter auxiliary networks for sample rectification \cite{shi2021online,mao2021adversarial,hwang2023aid,tsai2023test,yang2024adversarial}. They purify samples by reducing the loss of adversarial samples in self-supervised learning auxiliary tasks. Nonetheless,  prior studies \cite{athalye2018obfuscated,croce2022evaluating} have indicated that many adversarial rectification works \cite{samangouei2018defense,song2018pixeldefend,shi2021online} exhibit vulnerability to a stronger adaptive attack on BPDA. The rectification method based on diffusion has recently been shown to significantly decrease its robustness when evaluated using alternative gradients \cite{lee2023robust}. \cite {yang2024adversarial} augments the self-supervised learning sample rectification approach by integrating a variational autoencoder framework as an auxiliary task and exhibits robust defense capabilities against BPDA adaptive attacks, showcasing promising avenues for sample rectification.

Different from previous research, our objective is to leverage the unexplored LE prior implied in various adversarial samples under different attackers, and delve into the possibility of universal adversarial robustness against unseen/agnostic attacks. We propose a LE prior guided reverse-forward adversarial rectification technique and achieve efficient test-time defense against various attacks with different intensity.

\section{Method}
In this section, we introduce REAL, which aims to enhance the generalization ability of defenses against various attack methods by leveraging shared prior knowledge about entropy to rectify samples. Specifically, we propose a Max-Min entropy optimization scheme and an attack-aware weighting mechanism based on two low-entropy priors. Additionally, REAL can be integrated as a plugin with other sample rectification methods. We have designed a comprehensive modification algorithm that combines REAL with these methods. The overall framework of our method is shown in Fig. \ref{fig3}.

\subsection{Preliminary}
Our approach can seamlessly integrate with self-supervised learning-based sample rectification techniques, bolstering their robustness and defense against various attacks. Prior to delving into our methodology, we first introduce this type of sample rectification approach.

Adversarial rectification models can be summarized as the following paradigm: given a well-pretrained model with two branch structures, namely the main task and auxiliary task, where the main task represents the ultimate classification objective and the auxiliary task involves self-supervised objective such as data reconstruction \cite{feng2019unsupervised,tsai2023test}, variational inference \cite{yang2024adversarial}, rotation prediction \cite{gidaris2018unsupervised}, label consistency \cite{he2020momentum,chen2020simple} or self-supervised contrastive learning \cite{mao2021adversarial}. These two branches share the encoding part defined as $E$. For a given input $x$, the encoder outputs $z=E\left(x, \theta_{enc}\right)$. The classifier is represented as $C$ and outputs prediction $\hat{y}=C\left(z, \theta_{cls}\right)$. The auxiliary task is represented as $H\left(z, \theta_{aux}\right)$. During the training phase, the two branches conduct joint training with the training goal as:
\begin{equation}
\begin{aligned}
\mathcal{L}_{\text {train}}=&\min_\theta\{ \mathcal{L}_{c l s}((C \circ E)(x), y,\theta_{enc},\theta_{cls})\\ & +  \mathcal{L}_{\text {aux }}((H \circ E)(x),\theta_{enc},\theta_{aux})\}
\end{aligned}
\label{eq1}
\end{equation}
\textcolor{black}{where $\mathcal{L}_{c l s}$ denotes a cross-entropy loss function for classification, $\mathcal{L}_{aux}$ is the auxiliary self-supervised objective, for example, reconstruction Mean Squared Error (MSE) loss.}

Due to the joint representation learned by self-supervised task and classification task during training phase, it can be assumed that if the loss of samples in self-supervised tasks is small, they also perform well in classification task. Therefore, it is possible to utilize the different performances of adversarial and clean samples in the self-supervised task during testing phase to purify samples. In testing phase, the sample is rectified using the loss from auxiliary tasks, denoted as $\mathcal{L}_{\text {aux}}(x)$. The detailed computation is outlined as follows: 
\begin{equation}
	\min_\delta \mathcal{L}_{\text {aux}}\left((H \circ E)\left(x+\delta\right)\right),\quad s.t. \|\delta\| \leq \epsilon_{\text {pfy}},
	\label{eq2}
\end{equation}
where $\delta$ is the rectification amount that is applied to the sample, which is calculated using gradient optimization techniques. Additionally, $\epsilon_{\text {pfy}}$ is defined as the allowable budget for the adversarial rectification.
\begin{figure}[t]
\centering
\subfloat[]
    {
        \includegraphics[width=0.48\linewidth]{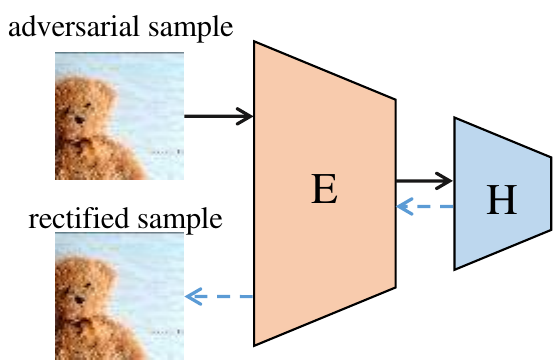}}
    \subfloat[]
    {
        \includegraphics[width=0.48\linewidth]{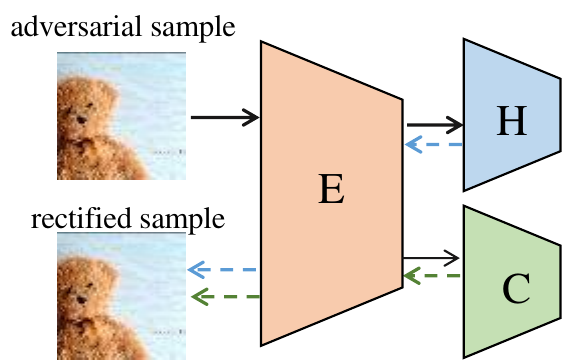}}
\caption{Difference in the basic architecture, where \textit{E}, \textit{H} and \textit{C} denote encoder, auxiliary task and classifier, respectively. (a) represents previous methods, which only uses self-supervised auxiliary task information for sample rectification, while (b) represents our method, which leverages self-supervised auxiliary tasks as well as predictive entropy information for sample rectification.}
\label{fig4}
\end{figure}
After obtaining the purified sample $x_{pfy}=x+\delta$, feed them into the model to obtain predictions as the final classification output  $y_{pre}=(C \circ E)(x_{pfy})$. 

Unlike the above traditional approaches that solely rely on the distinct performance of adversarial and clean samples in self-supervised auxiliary tasks for sample rectification, we aim to explore a different path by leveraging the different entropy characteristics of adversarial and clean samples to assist in sample rectification. To provide a more intuitive understanding of our method, we have drawn Fig. \ref{fig4}, which contrasts our approach with previous methods. The left side illustrates conventional methods, which rely on the alignment characteristics between self-supervised task and main classification task, assuming that the ability of the self-supervised task can also be translated into the main classification task. As a result, it only uses auxiliary loss for a single rectification, which often has the drawback of insufficient rectification and does not utilize information from the main classification task. The right side depicts our method, which incorporates a set of entropy losses during rectification process by utilizing the entropy properties of adversarial samples, thereby taking advantage of information from the main classification task. Through two aspects of alignment: self-supervised task representation alignment and entropy property alignment, we fully exploit the performance differences between adversarial and natural (clean) samples to rectify adversarial samples.

\subsection{Max-Min entropy optimization scheme}
In this paper, we delve into the properties of adversarial samples' prediction entropy. We have identified two key characteristics of adversarial samples in terms of predictive entropy that distinguish them from natural samples. Specifically, \textbf{Property 1} reveals that adversarial samples are often misclassified with low entropy and high confidence, contrasting with the distribution of natural samples, as shown in Fig. \ref{fig1}. Therefore, unlike the direct approach in test-time domain adaptation (DA) methods that achieve generalization in natural distribution samples by minimizing predictive entropy during testing, we need to design different robust generalization strategies tailored to the distinct entropy properties of adversarial samples. 

To address the issue of how to effectively utilize the entropy of adversarial samples, we propose a Max-Min entropy optimization strategy, which consists of two main steps. Firstly, considering this characteristic of adversarial samples, we need to mitigate their adversarial nature. To this end, we implement an entropy maximization strategy, combined with an auxiliary task loss, to adjust the adversarial samples, resulting in a mask sample represented as $x_{mask}=x_{adv}+\delta_{\text{mask}}$. The aim is to disrupt their inherent adversarial properties and introduce a masking effect and we achieve this by minimizing the masking loss $\mathcal{L}_{\text{mask}}$, reflected as follows: 
\begin{equation}
\begin{aligned}
\mathcal{L}_{\text{mask}} &= \min_\delta \{ \mathcal{L}_{\text{aux}}((H \circ E)(x_{\text{adv}}+\delta_{\text{mask}}))
\\& - \beta_{\max} \cdot \mathcal{L}_{\text{ent}}((C \circ E)(x_{\text{adv}}+\delta_{\text{mask}})) \}, s.t.\ \| \delta_{\text{mask}} \| \leq \epsilon_{\text{pfy}}
\end{aligned}
\label{eq3}
\end{equation}
where $\beta_{\max}$ is a trade-off parameter between two losses,  associated with attack strength and $\mathcal{L}_{\text{ent}}$ is represented by: 
\begin{equation}
\mathcal{L}_{\text{ent}}(x)=-\sum_{i=1}^n p_i \log \left(p_i\right)
\label{eq4}
\end{equation}
where $p_i$ represents the probability of predicting $x$ as class $i$, and $n$ is the total number of classes. 

\begin{figure}[t]
\centering
\includegraphics[width=\linewidth]{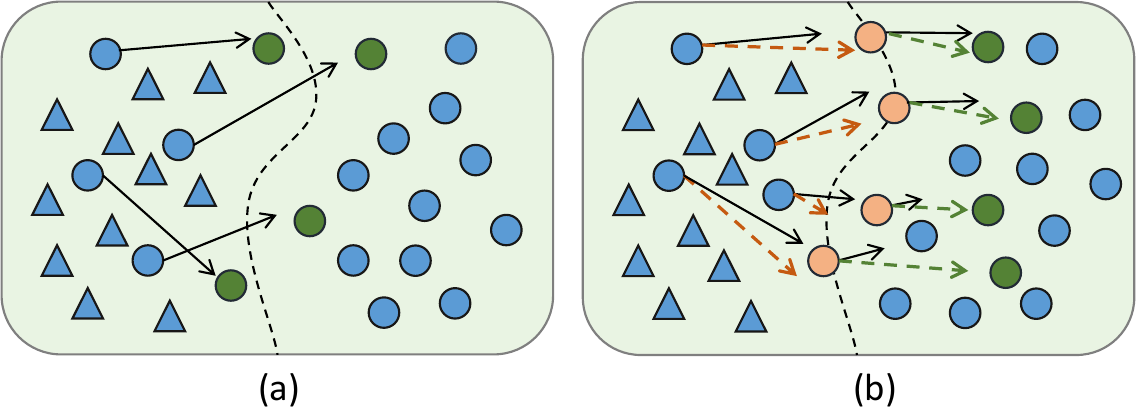}
\caption{(a) and (b) represent one-stage and two-stage rectification, respectively. Triangle and circular represent two classes. Among them, orange stands for mask samples and green represents final purified samples.}
\label{fig5}
\end{figure}
After the first stage (a.k.a. reverse rectification stage), we obtain the mask samples $x_{mask}$, believing that the adversarial characteristics of samples have been diminished. However, our ultimate goal is not only to eliminate the adversarial characteristics of samples but also to ensure that the samples are correctly classified, converging towards clean samples, that is, they can be correctly classified with high confidence. Therefore, we introduce the second stage (a.k.a. forward rectification stage) of enlightenment operations. In this step, we further incorporate entropy minimization to enlighten the samples, so as to achieve the final purified samples $x_{pfy}=x_{mask}+\delta_{\text{pfy}}$. The entropy minimization task is formulated as:
\begin{equation}
\begin{aligned}
\mathcal{L}_{\text{pfy}} &= \min_\delta \{ \mathcal{L}_{\text{aux}}((H \circ E)(x_{\text{mask}}+\delta_{\text{pfy}}))
\\& + \beta_{\min} \cdot \mathcal{L}_{\text{ent}}((C \circ E)(x_{\text{mask}}+\delta_{\text{pfy}})) \}, s.t.\ \| \delta_{\text{pfy}} \| \leq \epsilon_{\text{pfy}}
\end{aligned}
\label{eq5}
\end{equation}
where $\beta_{\min}$ is a trade-off parameter between the two losses. 

The conceptual framework of the two-stage process is illustrated in Fig. \ref{fig5}. In the first phase, by maximizing entropy, as shown by orange dashed line in this figure, we pull the adversarial samples with originally low entropy and high confidence misclassifications back towards the decision boundary, thereby disrupting the malicious adversarial perturbations. In the second phase, we further refine the samples through enlightenment processing by minimizing entropy, as shown by green dashed line in the figure. Compared with one-stage sample rectification method in Fig. \ref{fig5} (a), the two-stage rectification strategy that incorporates an entropy mechanism can more deeply explore the differences between adversarial and clean samples, thus achieving a more reliable purification of samples and correct malicious pixels.

Furthermore, the aforementioned two-stage Max-Min entropy optimization scheme can be viewed as an autonomous adversarial rectification procedure. During this self-adversarial rectification, the predictions for adversarial samples are engaged in a dynamic interplay between deterministic and uncertain outcomes. This process ultimately yields purified samples that satisfy the established criteria for rectification.

\subsection{Attack-aware weighting mechanism}
Regarding \textbf{Property 2} as shown in Fig. \ref{fig2}, it indicates the variations in the predicted entropy of adversarial samples generated by different attack methods. Specifically, most adversarial attack methods generate adversarial samples with a phenomenon of low-entropy misclassification, which is referred to as Property 1. Naturally, the degree of this phenomenon varies among different adversarial methods, but they all adhere to a negative correlation between entropy and attack strength. That is, as the attack strength increases (quantified by classification loss), the predictive entropy tends to decrease.

Given the diversity in the attack strength exhibited by adversarial samples crafted through various methods, a one-size-fits-all approach to their rectification is not flexible. For instance, in the first stage of rectification, it is impractical to uniformly maximize the entropy for an adversarial sample that has a higher attack strength and a low predictive entropy, as well as for a sample with a lower attack strength and a higher predictive entropy. This distinction necessitates a more nuanced and adaptive strategy for entropy maximization that takes into account the specific adversarial traits of each sample. Similarly, there also exists the same issue for the second stage of rectification. Therefore, we further propose an attack-aware weighting mechanism that considers the attack strength by introducing a dynamic rectification strategy with an adaptive parameter $\beta$.

Firstly, inspired by the relationship between predicted entropy and attack strength, we evaluate the attack strength of an adversarial sample by determining its normalized predicted entropy value, denoted as $\mathcal{V}_{ent}$, which is computed as:
\begin{equation}		
\mathcal{V}_{\text{ent}} = \mathcal{L}_{\text{ent}} / \log_2\left(N\right)
\label{eq6}	
\end{equation}
where \textit{N} is the number of categories. For adversarial samples with lower attack strength, i.e. higher $\mathcal{V}_{ent}$, it requires lighter entropy maximization. Therefore we design $\beta_{\max}$ to dynamically adjust the entropy maximization degree, as shown below:
\begin{equation}		
\beta_{\max} = \alpha \cdot \left(1 - \mathcal{V}_{\text{ent}}\right)^2, \quad  \mathcal{V}_{\text{ent}} \in [0,1]
\label{eq7}	
\end{equation}
where $\alpha$ represents a hyperparameter. As the predicted entropy $\mathcal{V}_{\text{ent}}$ decreases, the attack strength increases, leading to an increase in $\beta_{\max}$. By regulating $\beta_{\max}$, we apply a greater weight to entropy maximization for samples with stronger attack strength, thereby more effectively disrupting their adversarial characteristics. 

After the first stage, we obtain mask samples, but our expectation is to ultimately obtain purified samples that can approach clean samples, achieving high-confidence correct classification. To this end, we propose to implement the second stage: entropy minimization, which serves as a functional process of \textit{enlightenment}. This process is designed to guide the mask samples towards exhibiting traits that are more consistent with those of clean non-adversarial data. Refining these samples enables them to evolve from their original form to a state that aligns with the desired outcomes of the classification task. Similarly, due to the varying performance of the mask samples obtained in the first step, we also need to apply different degrees of entropy minimization. Since the adversarial characteristics of the mask samples have been eliminated, for samples with higher predictive entropy, that is, samples with uncertain predictions, we need to apply a greater weight for entropy minimization. We implement this by designing $\beta_{\min}$ as described in the following equation:
\begin{equation} 		
\beta_{\min} = \alpha \cdot \mathcal{V}_{\text{ent}}^2,\quad  \mathcal{V}_{\text{ent}} \in [0,1]
\label{eq8} 	
\end{equation}
where $\alpha$ represents a hyperparameter. As the predicted entropy $\mathcal{V}_{\text{ent}}$ increases, its distance from the ideal final output also increases. Therefore, we need to apply a greater degree of enlightenment processing. That is, greater weights regulated as $\beta_{\min}$ are assigned.
\begin{figure*}[t]
\centering
\includegraphics[width=\linewidth]{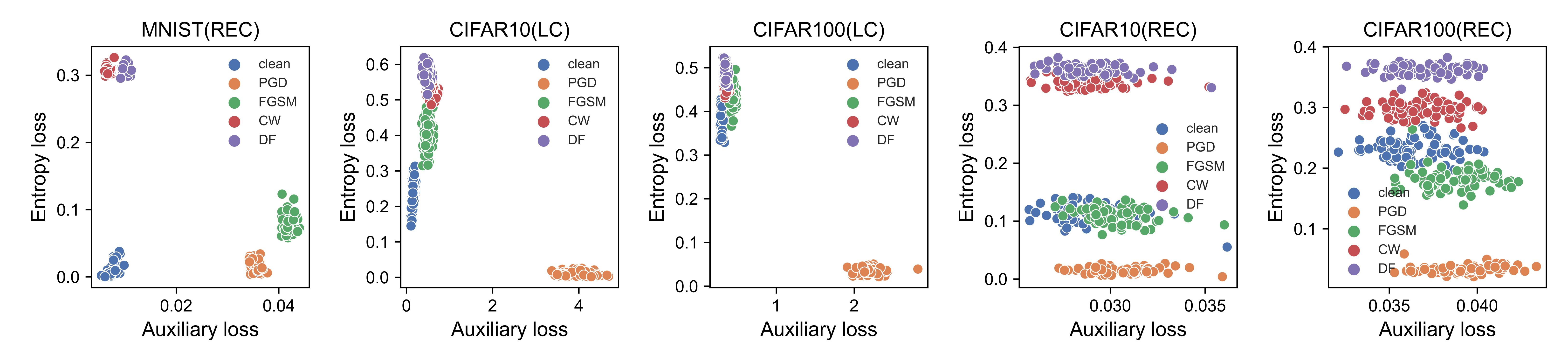}
\caption{Joint distributions of auxiliary loss and entropy loss. Clean samples (in blue) show different distribution from adversarial samples under various attacks. REC and LC respectively represent data reconstruction and label consistency auxiliary tasks.}
\label{fig6}
\end{figure*}
\subsection{Overall algorithm: multi-step optimization}
We provide a comprehensive algorithm structure by combining the above two components. In addition, while designing the final rectification algorithm, we also consider the trade-off between the accuracy of clean samples and adversarial samples, as well as the issue of insufficient rectification optimization. Therefore, we separately design a detection strategy based on loss joint distribution and a heuristic selection strategy (HSS). Algorithm \ref{algorithm} summarizes the procedure. 

\textbf{Detection strategy}: While our refinement algorithm has been successful in correcting adversarial samples, it is unavoidable that the accuracy of clean samples may decrease. To avert a precipitous drop in the precision of clean samples, we have incorporated a detection process to distinguish between clean and adversarial samples. Taking inspiration from \cite{tsai2023test}, we have plotted the joint distribution of auxiliary loss and predicted entropy across three datasets in Fig. \ref{fig6}, where each point on the graph represents the distribution within a batch. From the figure, it can be observed that there is a distinctive difference in the distribution between clean and adversarial samples. Additionally, we investigate the utilization of auxiliary loss for detection and observe that adding entropy loss can enhance sample detection. Specifically, setting both auxiliary loss thresholds, denoted as $aux^*$, and entropy loss thresholds, denoted as $ ent^*$, can act as a detection mechanism, effectively distinguishing between clean and adversarial samples. \textcolor{black}{The specific implementation is as follows. For a given input $x$, if it simultaneously satisfies the conditions of $\mathcal{L}_{aux}(x)<aux^* \text{ and } \mathcal{L}_{ent}(x)< ent^*$, the sample is output as the final rectified result. Otherwise, we perform REAL for sample rectification. The specific values of $aux^*$ and $ent^*$ are determined based on the statistical means of the auxiliary loss and predicted entropy of clean samples, respectively.} 

\textbf{Heuristic selection strategy}: Additionally, our method can be regarded as a form of self-adversarial rectification process. During the rectification, the final predictions of the samples oscillate between certain values and uncertain probabilistic forecasts, which, in a sense, increases the challenge of the rectification. Consequently, this may result in a single comprehensive rectification not being entirely sufficient. Inspired by multi-step attacks \cite{madry2017towards}, we adopt a multi-step iterative update design to increase the rounds of self-adversarial purification rectification. At the same time, to determine the number of self-purification rounds required for each sample, we have designed a heuristic selection strategy (HSS). This strategy dictates that when the purified sample meets a certain rectification termination condition, the sample is directly outputted; otherwise, it proceeds to the next rectification round. 

Here, we have established two termination conditions. The first condition is relatively more stringent, requiring that both the auxiliary loss $\mathcal{L}_{\text{aux}}(x_{\text{pfy}})$ and the predicted entropy loss $\mathcal{L}_{\text{ent}}(x_{\text{pfy}})$ of sample be less than the set threshold values to be directly outputted. That is, the rectification sample must align very well with the clean samples, or the sample itself must be a clean sample, to be considered for output. This condition can help detect clean samples that have not been detected. The second condition is more flexible, requiring that the auxiliary loss of sample be below the threshold. Additionally, if there is label reversal, referring to inconsistency between label predictions of purified sample and original sample, purified sample can be directly outputted:
\begin{equation} 		
LR =\left\{\begin{array}{l}True, \quad (C \circ E)(x_{pfy})\neq (C \circ E)(x) \\
False, \quad \text {otherwise}
\end{array}\right.
\label{eq9} 	
\end{equation}
where LR means label reversal.
Describe the overall cut-off conditions as follows:
\begin{equation}
\begin{aligned}
\left\{\begin{array}{l}
\mathcal{L}_{\text{aux}}(x_{\text{pfy}})<aux^*\text { and }  \mathcal{L}_{\text{ent}}(x_{\text{pfy}})<ent^*, \\
\mathcal{L}_{\text{aux}}(x_{\text{pfy}})<aux^*\text { and } LR
\end{array}\right.
\end{aligned}
\label{eq10}
\end{equation}


when $x_{\text{pfy}}$ meets one of the conditions in Eq. \ref{eq10}, rectification is completed. In summary, the overall rectification process is as follows: Initially, a test sample is detected. If the sample's auxiliary loss and predicted entropy loss are below thresholds, it is considered a clean sample without needing rectification. Otherwise, the rectification process is initiated. Rectification is a two-stage process. In the first stage, a masking treatment is applied, using ${L}_{mask}$ to calculate rectification amount, resulting in ${x}_{mask}$. In the second stage, sample is subjected to enlightenment treatment, using ${L}_{pfy}$ to calculate rectification amount, resulting in ${x}_{pfy}$. At this point, it is determined whether the sample meets rectification termination conditions. If conditions are met, the corrected sample is outputted; if not, this sample proceeds to the next round of rectification.

\begin{algorithm}[t]
\caption{Rectify Adversarial Sample via Max-Min Entropy Optimization for Test-Time Defense}
\label{alg:Framwork}
\begin{algorithmic}[1]
    \Require $x$: Test sample, $aux^*$: Auxiliary loss detection threshold, $ent^*$: Information entropy detection threshold, $T$: the number of rectification steps, $\gamma$: Purification step size, $R$: the maximum number of rectification rounds.
    \Ensure $x_{\text{pfy}}$: Purified sample.
    \If {$\mathcal{L}_{aux}(x) < aux^* \text{ and } \mathcal{L}_{\text{ent}}(x) < ent^*$} 
    \State Return $x_{\text{pfy}} = x$ 
    \Else
    \State Initialize $\delta = 0$, $x_{\text{pfy}} = x$, $rounds = 1$
    \Repeat
    \State \textbf{Stage 1: reverse rectification stage}
    
    $\bullet$ Compute $\mathcal{V}_{\text{ent}}(x_{\text{pfy}})$ and $\beta_{\max}$ using Eq. \ref{eq4}, \ref{eq6} and \ref{eq7};
    
    \textcolor[rgb]{0.13,0.73,0.87}{\# Calculate predicted entropy of sample and maximum entropy weight for the first stage.}

    $\bullet$ Compute $\mathcal{L}_{\text{mask}}$ according to Eq. \ref{eq3};

    \textcolor[rgb]{0.13,0.73,0.87}{\# Calculate the first stage rectification loss.}
  
    $\bullet$ Compute $\delta_{\text{mask}}=\delta_{\text{mask}} - \gamma \cdot \text{sign}\left(\nabla_{x} \mathcal{L}_{\text{mask}}\right)$ and obtain $x_{\text{mask}}$=$x$ + $\delta_{\text{mask}}$
    
    \textcolor[rgb]{0.13,0.73,0.87}{\# Perform a single step gradient descent.}

    \State \textbf{Stage 2: forward rectification stage}
    
    $\bullet$ Compute $\beta_{\min}$ using Eq. \ref{eq8};
    
    \textcolor[rgb]{0.13,0.73,0.87}{\# Calculate minimum entropy weight.}

    $\bullet$ Compute $\mathcal{L}_{\text{pfy}}$ according to Eq. \ref{eq5};

    \textcolor[rgb]{0.13,0.73,0.87}{\# Calculate the second stage rectification loss.}

    $\bullet$ Compute $\delta_{\text{pfy}}=\delta_{\text{pfy}} - \gamma \cdot \text{sign}\left(\nabla_{x} \mathcal{L}_{\text{pfy}}\right)$ and obtain $x_{\text{pfy}}$=$x_{\text{mask}}$ + $\delta_{\text{pfy}}$

    \textcolor[rgb]{0.13,0.73,0.87}{\# Perform a single step gradient descent.}

    \State$rounds$ += 1
    \Until{$x_{\text{pfy}}$ meets condition in Eq. \ref{eq10}  or $rounds$ $\geq$ $R$}
    Return $x_{\text{pfy}}$
    \EndIf
\end{algorithmic}
\label{algorithm}
\end{algorithm}

\section{Experiments}
\subsection{The selection of datasets and backbone}
We validate our method on four commonly used classification datasets, MNIST \cite{1998Gradient}, CIFAR10, CIFAR100 \cite{krizhevsky2009learning}, and TinyImageNet. 

$\bullet$ \textbf{MNIST} is a dataset of handwritten digits commonly used for training various image processing systems. It contains 60,000 training images and 10,000 test images, each of 28x28 pixels in grayscale, categorized into 10 classes (digits 0 through 9). 

$\bullet$ \textbf{CIFAR10} comprises 60,000 color images of 32x32 in 10 different classes, with 50,000 training images and 10,000 test images. The categories include airplanes, cars, birds, cats, deer, dogs, frogs, horses, ships, and trucks.

$\bullet$ \textbf{CIFAR100} is similar to CIFAR10 but contains 100 classes with 600 images per class, divided into 50,000 training images and 10,000 test images. Each image is also 32x32 pixels in RGB. 

$\bullet$ \textbf{TinyImageNet} is a subset of the ImageNet dataset, containing 200 classes with 500 training images, 50 validation images, and 50 test images per class. Each image is resized to 64x64 pixels in RGB.

We use different backbone architectures for each dataset: a fully-connected network (FCN) and a convolutional neural network (CNN) \cite{shi2021online} architectures for MNIST, and ResNet18 \cite{he2016deep}, WideResNet28-10 \cite{zagoruyko2016wide} architectures for CIFAR10/100 and TinyImageNet, respectively. In addition, we validate PreresNet18 \cite{he2016identity} on the CIFAR10. For fair comparison, we adopt the same training parameters as \cite{shi2021online} and \cite{mao2021adversarial}.

\subsection{The selection of self-supervised tasks}
The proposed REAL can work as a plugged-played block with existing models that exploits self-supervised tasks for sample rectification. Self-supervised auxiliary tasks have various designs, and in our specific experiments, we explore settings such as data reconstruction, label consistency, and self-supervised contrastive learning tasks. The data reconstruction task (REC) is designed with an auxiliary network as the decoder, encoding hidden features into images to make the reconstructed image similar to the original input pixels. \textcolor{black}{The label consistency task (LC) does not require additional network structure design, which has low computational complexity. It requires transformation (e.g., rotation), such that the transformed image is enabled to be consistent with the original image prediction. In contrast, self-supervised contrastive learning tasks aim to optimize the relationships between enhanced sample pairs. Through data augmentation, positive and negative pairs are generated, with the objective of maximizing the consistency between positive pairs while minimizing the consistency between negative pairs.}
In practice, we use the reconstruction task for MNIST due to image transformation issues in other self-supervised tasks such as rotation. For CIFAR10/100, we assess both reconstruction and label consistency tasks, and validate the self-supervised contrastive learning auxiliary task on CIFAR10. For TinyImageNet, we assess label consistency tasks.
\begin{table*}[t]
\setlength{\tabcolsep}{3mm}
\caption{Comparisons of our method with previous defense models on MNIST, CIFAR10, CIFAR100, and TinyImageNet datasets. \textcolor{black}{For MNIST, we test FCN as backbone and for CIFAR10/100, we test ResNet18 as backbone.}}
\begin{center}
\begin{tabular}{|c|c|ccccccccc|}
\hline
Dataset      & Method    & Natural    & FGSM  & PGD     & FGSM-T & PGD-T    & CW    & DF   & AA  & Worst                                    
\\ \hline
\multirow{6}{*}{MNIST}       
& None  & \ul{\textbf{98.10}} & 16.87 & 0.49  & 15.67    &10.13  &  0.01 &  1.40 & 0.00  & 0.00  \\
& FGSM-AT &  79.76 &  \textbf{80.57}&  2.95 &  56.25 &  10.65   & 6.22  & 17.24  &  0.00  & 0.00 \\
& PGD-AT  &  76.82 &  60.70&  57.07 &76.19 &76.46 &31.68 &13.82 &46.61 & 13.82 
\\
& Defense-GAN  &95.84 & 79.30 &\textbf{84.10} &- &- &\textbf{95.07} & {\textbf{95.29}} & -  & \textbf{79.30}   
\\
& SOAP (REC) & \textbf{97.56} &66.85 &61.88 & \textbf{77.25}  & \textbf{85.95}     &86.81       & 87.02  &\textbf{55.62}     &55.62
\\
& SOAP+Ours (REC)    & 97.39  &  \ul{\textbf{95.85}} & \ul {\textbf{96.82}} & \ul{\textbf{92.67}} & \ul {\textbf{94.14}} & \ul{\textbf{98.28}} &\ul {\textbf{97.60}} & \ul{\textbf{91.63}} &\ul{\textbf{91.63}}
\\[2pt]\hline
\multirow{14}{*}{CIFAR10} 
& None  
&\ul{\textbf{90.54}} & 15.42  & 0.00  & 19.06 & 10.00  & 0.00 & 6.26  & 0.00 & 0.00
\\
& FGSM-AT & 72.73  & 44.16 & 37.40   & 61.75   & 10.09  & 2.69   & 24.58  & 0.00 & 0.00                                     \\
& PGD-AT & 74.23 & 47.43  & 42.11  & \ul{\textbf{69.60}} & 68.99 & 3.14 & 25.84  &43.30    & 3.14    \\
& TRADES & 83.22   & \textbf{58.51} & 54.97  & - & - & 71.62   & -   & 48.97        & 48.97         \\
& MART  & 82.14 & \ul{\textbf{59.57}} & \textbf{55.39}  & -   & -  & 74.30 & -  &56.24 &\textbf{55.39} \\
 & Pixel-Defend & 79.00 & 39.85 & 29.89   & -   & -   & \textbf{76.47}  & \textbf{76.89}& -           &29.89     \\
& SOAP (REC)    & 78.10& 24.29 & 17.29& 44.28  & 46.01& 66.50 & 65.97  & 34.15        &17.29      \\
& SOAP+Ours (REC)    & 67.77  & 37.11  & 31.52 & 50.64 & 53.64 & 67.42& 65.52  & 36.20    &31.52  \\
& SOAP (LC)     & \textbf{84.07} & 51.02  & 51.42 & 63.23    & \textbf{72.38}              & 73.95    &74.79 & \textbf{70.37}    &51.02   
\\
& SOAP+Ours (LC)     & 78.82  & 58.29 & \ul{\textbf{62.43}}    & \textbf{66.44}   & \ul{\textbf{75.90}}            & \ul{\textbf{85.80}}   & \ul{\textbf{82.01}}   & \ul{\textbf{70.48}}       &\ul{\textbf{58.29}}   
\\ [2pt]\hline 
\multirow{9}{*}{CIFAR100}     
& None  & \ul{\textbf{65.56}}   & 3.81  & 0.01  & 6.42 & 1.13  & 0.00  & 12.30 & 0.00  & 0.00 \\
& FGSM-AT & 44.35 & 20.30 & 17.41  & 36.96  & 2.74  & 4.23   & 18.15 & 0.00 & 0.00                                    \\
& PGD-AT & 42.15 & 21.92 & 20.04 & \ul{\textbf{42.59}} & 42.88 & 3.57 & 17.90 & 28.82  & 3.57 \\
& TRADES & 53.93  & 30.22  & 28.06  & -    & -   & 35.35  & -  & 23.01    & 23.01                                \\
& MART  & \textbf{55.52}   & \ul{\textbf{30.82}} & \textbf{28.38}  & \textbf{-}         & -  & 35.57    & -    & 23.33 &\textbf{23.33}                                  
\\
& SOAP (REC) & 52.46  & 9.34  & 6.74    & 10.19  & 16.38 & 40.36  & 41.19 & 2.67     & 2.67      \\
& SOAP+Ours (REC)    & 37.14& 14.51 & 10.80 & 10.15   & 15.31  & 41.51& 39.40  & 6.15   & 6.15   \\
& SOAP (LC)  & 52.91 & 22.93 & 27.55 & 39.16  & \textbf{45.50} & \textbf{50.26}   & \textbf{50.57}   & \ul{\textbf{45.11}} & 22.93
\\
& SOAP+Ours (LC)     & 44.27   & \textbf{30.49}  & \ul{\textbf{35.54}}   &\textbf{40.95}  & \ul{\textbf{46.20}}   & \ul{\textbf{55.26}}   & \ul{\textbf{51.60}} &\textbf{43.69} & \ul{\textbf{30.49}} \\ [2pt]\hline

\multirow{5}{*}{TinyImagenet} & None         & \ul{\textbf{51.01}}                     & 2.00    & 1.58      &  2.15  & 0.53 & 0.00& 11.85   &  0.00    &  0.00    
\\
& FGSM-AT  & 29.00  & 13.45 & 12.29 &  \textbf{27.44} & 27.55  & 8.42 & 18.19   & 16.23  &  \textbf{8.42}\\
& PGD-AT & 28.49 & \ul{\textbf{13.70}}  & \textbf{12.98} &\ul{\textbf{27.99}}  &  28.13   & 8.16  & 17.96   &  \textbf{17.90}    &8.16         
  \\
& SOAP (LC) & \textbf{41.97}& 7.97& 5.69  &  16.19 &  \ul{\textbf{30.89}}& \textbf{40.29}     & \textbf{40.65} & 12.89 &5.69 
 \\
& SOAP+Ours (LC)     & 30.90 &  \ul{\textbf{29.90}} & \ul{\textbf{13.35} }&22.34 & \textbf{29.87}       & \ul{\textbf{41.03}}  & \ul{\textbf{42.62}}  &  \ul{\textbf{21.10}}    &\ul{\textbf{13.35}}
\\ [2pt]\hline
\end{tabular}
\end{center}
\label{table1}
\end{table*}
\subsection{The selection of attack methods}
We conduct experiments on several common attacks, including FGSM, PGD, CW, DeepFool (DF) and Autoattack (AA). For these attacks, we adopt common attack parameter settings. Except for the CW and DF attacks, the maximum perturbation strength is set to $L_{\infty}$ bounded with ${\epsilon}$ = 0.3 for MNIST, and $L_{\infty}$ bounded with ${\epsilon}$ = 8/255 for CIFAR10/100, TinyImageNet. For AA, we choose the standard option. Additionally, the number of attack steps for PGD and PGD-target is set to 40 with a step size of 0.01 for MNIST, and 20 steps with a step size of 2/255 for CIFAR10/100, TinyImageNet. \textcolor{black}{For the CW and DF attacks, which are optimization-based, we set the maximum perturbation strength to $L_2$ bounded with ${\epsilon}$ = 4.} At the same time, further experimental analysis is conducted on adaptive attacks (i.e. attacking with all defense strategies known), ultimately demonstrating the effectiveness of our method. We have investigated two types of adaptive attacks. The first type is the commonly used multi-target attack, which includes auxiliary rectification loss and classification loss, as utilized in previous works \cite{shi2021online,mao2021adversarial}. The second type is the stronger BPDA attack proposed by \cite{croce2022evaluating}. To ensure a fair comparison, we employ identical attack parameter configurations as described in \cite{shi2021online,mao2021adversarial}. 
\subsection{The selection of defense method parameters}
Firstly, in our method, in order to prevent a serious decrease in the accuracy of clean samples, we attempt to set a joint threshold of auxiliary loss and entropy loss for detecting clean and adversarial samples, as shown in Fig. \ref{fig6}. We approximate the detection threshold based on statistical data obtained from clean samples. However, the detection methods have different adaptability to different auxiliary tasks. Due to the complexity of reconstruction tasks on CIFAR10/100 datasets, the distribution of auxiliary losses is mixed, making detection challenging. Therefore, we choose not to employ threshold detection for data reconstruction tasks. In addition, in self-supervised contrastive learning auxiliary task, in order to facilitate the combination of methods, we omit the detection stage. Secondly, our strategy employs a two-stage rectification approach. To reduce computational overhead, we set the number of iterations $T$ in each stage to 3 for four datasets. Besides we set an iteration step $\gamma$ of 0.1 for MNIST and 4/255 for CIFAR10/100. During rectification phase, we set the trade-off parameter $\alpha$ to 0.25 and the maximum number of rectification rounds $R$ to 5. 
\begin{table*}[t]
\setlength{\tabcolsep}{3mm}
\caption{Comparisons of our method with previous defense models on MNIST, CIFAR10, CIFAR100, and TinyImageNet datasets. For MNIST, we select CNN as backbone and for CIFAR10/100, we select WideresNet28-10 as backbone.}
\begin{center}
\begin{tabular}{|c|c|ccccccccc|}
\hline
Dataset   & Method  & Natural  & FGSM & PGD  & FGSM-T & PGD-T & CW & DF& AA & Worst \\ \hline
\multirow{6}{*}{MNIST}       
& None  & \ul{\textbf{99.15}} & 1.49  & 0.00 & 10.27 & 12.73 & 0.00& 0.00 & 0.00 & 0.00  
\\
 & FGSM-AT      
 & 98.78  &\ul{\textbf{99.50}} &33.70 & 50.13 & \ul{\textbf{98.25}} & 0.02 & 6.16 & 0.00 & 0.00 \\
& PGD-AT       
& 98.97  & \textbf{96.38}  & \ul{\textbf{93.22}} & \ul{\textbf{98.13}} &\textbf{97.35}  & 90.31 & 75.55  & \ul{\textbf{88.26}} & 75.55
\\
& Defense-GAN  & 95.92  & 90.30 & 91.93  & -  & -  &\textbf{95.82} & \textbf{95.68} & -  &  \ul{\textbf{90.30}}\\
& SOAP (REC)  & \textbf{99.04} & 87.78 & 84.82 & 92.66 & 88.70& 74.61 & 81.27 & 67.98   & 67.98
\\
& SOAP+Ours (REC) & 98.85  & 93.97& \textbf{92.38} & \textbf{96.69}& 93.93
& \ul{\textbf{98.05}} & \ul{\textbf{96.42}} & \textbf{84.90}    & \textbf{84.90}   
\\ [2pt]\hline
\multirow{12}{*}{CIFAR10}      
& None         
& \ul{\textbf{95.13}} & 14.82 & 0.00 & 28.27& 10.00 & 0.00 & 3.28 & 0.01 & 0.00    \\
& FGSM-AT      
& 72.20 & \ul{\textbf{91.63}} & 0.01 & \ul{\textbf{80.23}} & 10.20 & 0.00& 14.41 & 0.40 & 0.00\\
& PGD-AT   
& 85.92  & 51.58 & 41.50 & 70.50& 72.21 & 2.06 & 24.08 & \textbf{65.73}    & 2.06  
\\
& Pixel-Defend & 83.68& 41.37  & 39.00  & - & - & 79.30 & \ul{\textbf{79.61}} & -  & 39.00      \\
& SOAP (REC)    & 76.67& 32.38 & 23.46 & 53.35 & 58.15& 62.07 & 64.61 & 22.40        & 22.40        \\
& SOAP+Ours (REC)  & 72.48  & 41.60 &35.32 &56.30 &57.80  &69.47 &\textbf{67.49} & 39.85  & 35.32\\
& SOAP (LC) & \textbf{91.89}  & 64.83 & \textbf{53.58}& 71.98 & \textbf{78.95} & \textbf{80.33} & 60.56 & 64.33   &\textbf{53.58}
\\
& SOAP+Ours (LC) & 91.02& \textbf{65.78}& \ul{\textbf{58.55}} & \textbf{72.31}
& \ul{\textbf{79.02}}  & \ul{\textbf{83.54}} & 59.87 & \ul{\textbf{66.06}} &\ul{\textbf{58.55}}      
\\ [2pt]\hline
\multirow{7}{*}{CIFAR100} & None 
& \ul{\textbf{78.16}} & 13.76  & 0.06& 10.49 & 1.20 & 0.01 & 9.05  & 0.00   & 0.00 \\
& FGSM-AT      
& 46.45 & \ul{\textbf{88.24}} & 0.15  & \textbf{49.68}  & 1.25 & 0.00  & 13.40  & 0.01    & 0.00   \\
& PGD-AT  &\textbf{62.71}& 28.15 &21.34 &\ul{\textbf{51.23}} &42.76 &0.65  &16.57 & 34.02  & 0.65  \\
& SOAP (REC) & 53.59 & 21.36  & 17.77 & 30.40 & 37.65 & 42.73  & 42.87& 23.65        &17.77        \\
& SOAP+Ours (REC)& 48.46   & 23.78 & 20.04 & 30.50  & 35.50& 46.16  & 45.02& 25.10   &20.04      \\
& SOAP (LC) & \textbf{61.01} &31.40 &\textbf{37.53} & 37.22 &48.67 &\ul{\textbf{56.09}} &\textbf{53.79} & \ul{\textbf{51.56}} &\textbf{31.40}
\\
& SOAP+Ours (LC) & 56.57 & \textbf{31.95}& \ul{ \textbf{39.25}} & 38.36& \ul{\textbf{49.05}}    & \textbf{54.65}       & \ul{\textbf{54.75}}       & \textbf{47.13}  &\ul{\textbf{31.69}}    
\\ [2pt]\hline
\multirow{5}{*}{TinyImagenet} & None  
& \ul{\textbf{65.89}} & 8.96   & 0.26  & 4.14   & 0.77& 0.00& 9.66 & 0.10  & 0.00  \\
& FGSM-AT      
& 50.78  &\ul{\textbf{24.00}} & 19.97 & \ul{\textbf{49.20}} & 0.81& 10.25 & 24.65& 0.00 & 0.00 \\
& PGD-AT & 49.45 & 22.14 & 20.18 & \textbf{31.09}  & \ul{\textbf{41.91}} & 8.95   & 26.87  & 21.36  & 8.95        
\\
& SOAP (LC) & 47.23& 21.20 & \textbf{25.09} & 18.40  & 35.79  & \textbf{44.12}  & \textbf{42.20} & \textbf{29.86}   &\textbf{18.40}  
\\
& SOAP+Ours (LC)  & \textbf{50.80}  & \textbf{22.55} & \ul{\textbf{30.95}} & 23.86  & \textbf{38.30}  & \ul{\textbf{44.45}} & \ul{\textbf{42.88}} & \ul{\textbf{35.90}} &\ul{\textbf{22.55}}
\\ [2pt]\hline
\end{tabular}
\end{center}
\label{table2}
\end{table*}
\subsection{Main results}
We compare adversarial training, represented by AT (FGSM) \cite{goodfellow2014explaining} and AT (PGD) \cite{madry2017towards}, TRADES \cite{zhang2019theoretically}, MART \cite{wang2019improving}, as well as several sample modification methods, including Pixel Defend \cite{song2018pixeldefend}, Defense GAN \cite{samangouei2018defense}, SOAP \cite{shi2021online}. Additionally, by integrating our method with SOAP, we have demonstrated that our approach can serve as a plug-in to further enhance the defensive generalization capabilities of existing methods. To validate our approach's effectiveness, we perform comparative experiments across diverse backbone architectures and different combination methods designs. Firstly, Table \ref{table1} and \ref{table2} show the performance of different backbones. Specifically, Table \ref{table1} shows ResNet18 and FCN backbones for CIFAR10/100, TinyImageNet and MNIST datasets. Table \ref{table2} presents WideResNet28-10 backbone for the same datasets, excluding MNIST's CNN configuration. Secondly, when combined with SOAP, we have implemented a variety of self-supervised tasks for each dataset. The brackets in Table \ref{table1}, \ref{table2} denote different auxiliary tasks are utilized, where LC stands for Label Consistency auxiliary task and REC signifies the Data Reconstruction task. Furthermore, the optimal values are underlined, whereas the suboptimal values are in bold.

For CIFAR10/100, using label consistency task as auxiliary network can achieve the optimal effect in our method and the overall results are superior to traditional adversarial training methods. \textcolor{black}{Compared with SOAP (LC) \cite{shi2021online}, adding our method can further improve the accuracy under most attacks, by 7\%/7\% under FGSM attack}. In data reconstruction task, although the optimal accuracy is not achieved, it can still be observed that the effect is significantly improved after adding our method. To evaluate efficacy of our approach, we conduct comparative experiments on a larger dataset, TinyImagenet, and implement the setting of label consistency auxiliary task, which shows good results on both CIFAR10/100. In all datasets, we report robustness of the worst-case scenario. It can be seen that our method can enhance the robustness in the worst-case scenario, indicating an improvement in the generalization of adversarial robustness.

Additionaly, as shown in Table \ref{table1}, \ref{table2}, the integration of our method notably enhances defense against FGSM and PGD attacks. 
\begin{table*}[t]
\setlength{\tabcolsep}{3mm}
\caption{Comparative experiments on the CIFAR10 dataset by plugging and playing REAL in other model.}
\begin{center}
\begin{tabular}{|c|c|ccccccc|}
    \hline
Backbone   & Method  & Natural  & PGD50   & PGD200    & CW200    & BIM200   & AA     & Worst  
\\ \hline
& No defense   & \textbf{92.28} & 0.00   & 0.00  & 0.00  & 0.00  & 0.00  & 0.00              \\
& Mao et al. \cite{mao2021adversarial}  & 74.94  & 16.01 & 18.05 & 17.02  & 18.53    & 18.37    & 16.01
\\
&  Mao et al. \cite{mao2021adversarial}+Ours & 64.92 & \textbf{29.82}  & \textbf{30.80} & \textbf{30.66}        & \textbf{31.29}   & \textbf{31.56}    & \textbf{29.82}   
\\
\multirow{-4}{*}{PreResNet18} & $\Delta$ & \color[HTML]{00FF00}{-10.02} & {\color[HTML]{FE0000} +13.81} & {\color[HTML]{FE0000} +12.75} & {\color[HTML]{FE0000} +13.19} & {\color[HTML]{FE0000} +13.64} & {\color[HTML]{FE0000} +12.76} & {\color[HTML]{FE0000} +13.81} \\ \hline
\end{tabular}
\end{center}
\label{table3}
\end{table*}
For all four datasets, our method demonstrates significant enhancement under PGD and FGSM attacks compared with other attacks (e.g., CW and DF). We also provide an explanation for this phenomenon. As shown in Fig. \ref{fig6}, under PGD and FGSM attacks, adversarial samples exhibit high-confidence misclassification and significantly different from the entropy distribution of clean samples. Consequently, upon integrating the Max-Min entropy self-adversarial rectification strategy, notable improvements in performance can be observed. In contrast, CW and DF operate on different principles and are designed to introduce subtle perturbations leading to misclassifications, resulting in a smaller difference between the entropy distributions of adversarial and clean samples. \textcolor{black}{In this regard, our method can still achieve some improvement effects, for example, it can improve by nearly 10\% on CIFAR10 (ResNet18) as shown in Table \ref{table1}.} However, we also observe on CIFAR100, our method is slightly inferior to SOAP at defending against CW and DF. The reason may be due to the poor detection results on these attack methods. As shown in Fig. \ref{fig6}, for CIFAR100, the distribution of clean samples and adversarial samples generated by CW and DF are mixed together. Therefore, it difficult to detect them by setting thresholds and most adversarial samples are incorrectly identified as clean samples, leading to insufficient optimization. Hence, in future research, the detection method for purified samples can be optimized for better sample rectification.

In addition, we add a set of contrastive learning task settings to the CIFAR10 dataset for validation, as shown in Table \ref{table3}. Our proposed insertion component aims to enhance the defense effectiveness of self-supervised sample rectification methods without relying on adversarial training. Hence, in this setup, we employ non-adversarial training models as the benchmark for comparison, by selecting PreResNet18 as the backbone. Initially, we evaluate the rectification performance of Mao's method \cite{mao2021adversarial} using self-supervised contrastive loss on the standard training model. Subsequently, we enhance this approach by incorporating the proposed two-stage self-adversarial entropy rectification mechanism (i.e. reverse rectification and forward rectification). We set the rectification round $R$ to 1 and omit the detection stage for the sake of method insertion convenience. The experimental findings in Table \ref{table3} indicate that integrating the adversarial entropy rectification mechanism further improves the robustness of the baseline method. The effect of our REAL is demonstrated.
\begin{table*}[t]
\centering
\setlength{\tabcolsep}{2.5mm}
\caption{Ablation results on CIFAR10 for the auxiliary self-supervised loss ($\mathcal{L}_{aux}$), Max-Min entropy optimization scheme (max-min), heuristic selection strategy (HSS) and attack-aware weighting mechanism ($\beta$), respectively.}
\begin{center}
\begin{tabular}{|c|cccc|ccccccccc|}
\hline
 \multirow{2}{*}{Auxilary} & \multicolumn{4}{c|}{Method} 
 & \multirow{2}{*}{Natural} & \multirow{2}{*}{FGSM} & \multirow{2}{*}{PGD} & \multirow{2}{*}{FGSM-T} & \multirow{2}{*}{PGD-T} & \multirow{2}{*}{CW} & \multirow{2}{*}{DF} & \multirow{2}{*}{AA} & \multirow{2}{*}{Worst}
\\ \cline{2-5}
& $\mathcal{L}_{\text{aux}}$ & max-min & HSS & $\beta$ 
&   &  &    &   &  &  & & & \multicolumn{1}{l|}{}
\\ \hline
\multirow{5}{*}{REC}  
&$\times$  &$\times$ &$\times$ &$\times$ & \textbf{83.24} & 12.88  & 1.59  &28.75 &15.33 &0.00& 10.31 & 0.89 & 0.00     \\ 
&$\times$ & \checkmark &$\times$  &$\times$  & 41.69 & 20.12 &19.17 &28.90 &39.78 & 36.18&50.69 &27.06 &19.17
\\
 &\checkmark &$\times$ &$\times$ &$\times$ & 78.10 &24.29  &17.29 &44.28 &46.01 &66.50 &65.97 &34.15 & 17.29
 \\
& \checkmark &\checkmark &$\times$ &$\times$ & 77.10 & 22.75 & 12.58  &45.72 &41.33& 66.75 &\textbf{66.35}     & 35.44 & 12.58
\\
&\checkmark &\checkmark &\checkmark &$\times$ & 67.20  & 36.89 & \textbf{31.84} & 49.76& 53.57 & 67.33  & 65.35& 36.01 & \textbf{31.84}
\\
&  \checkmark &  \checkmark   &  \checkmark&  \checkmark  & 67.77                   & \textbf{37.11}  & 31.52 & \textbf{50.64} & \textbf{53.64} & \textbf{67.42}  & 65.52            & \textbf{36.20}    & 31.52
\\ \hline
\multirow{5}{*}{LC}
&$\times$ &$\times$ &$\times$   &$\times$  & \textbf{86.42} & 22.04& 0.15 & 34.60& 12.85 & 0.00 & 8.62& 0.03  & 0.00
\\ 
&$\times$ &\checkmark &$\times$ &$\times$ & 44.70  & 21.00 & 23.54 & 27.62& 29.84&36.98&35.35& 29.73 &21.00
\\
&\checkmark  &$\times$ &$\times$ &$\times$ & 84.07  &51.02 & 51.42&63.23 & 72.38 &73.95 &74.79 &70.37 &51.02
\\
&\checkmark &\checkmark  &$\times$ &$\times$ & 84.16 &55.09 &54.44 & 63.91& 75.42& 81.27& 80.36  & 62.28 &54.44
\\
&\checkmark &\checkmark &\checkmark &$\times$ & 77.79 &58.14 &61.20 &65.31& 75.71&84.40 & 81.20 & 69.10 &58.14
\\
&\checkmark &\checkmark  &\checkmark  &\checkmark  & 78.82 & \textbf{58.29} & \textbf{62.43} & \textbf{66.44} & \textbf{75.90} & \textbf{85.80} & \textbf{82.01}  & \textbf{70.48} &\textbf{58.29}
\\ \hline
\end{tabular}
\end{center}
\label{table4}
\end{table*}
\subsection{\textcolor{black}{Analysis of Detection Strategy}}
\textcolor{black}{During the defense phase, we propose a novel detection strategy to alleviate the trade-off between clean sample and adversarial sample accuracy. To comprehensively evaluate our detection method, we compare it with LID\cite{ma2018characterizing}, Mahalanobis\cite{lee2018simple}, LiBRe\cite{deng2021libre} and EPS-AD\cite{zhang2023detecting} in terms of both principles and experimental performance.
From the perspective of detection principles, LID, Mahalanobis, LiBRe and EPS-AD all rely on specific detection models and additional optimization process. Among them, LiBRe has slower training and EPS-AD relies on diffusion models, increasing computational cost.
In contrast, our method does not require extra optimization steps or complex models, offering high efficiency. Additionally, although LID, Mahalanobis, and EPS-AD improve generalization to unknown attacks, they still rely on adversarial samples for training. While our method uses only clean sample statistics, which better aligns with real-world detection requirements.
In terms of experimental evaluation, We evaluate the detection accuracy of different attack methods on CIFAR10 using ResNet18 as the backbone. To test generalization to unknown attacks, we restrict the training sets of LID, Mahalanobis and EPS-AD to FGSM attack only. As shown in Table \ref{table5}, our method achieves a detection accuracy of over 80\% across various attacks. While the overall accuracy is slightly lower than other methods, this difference is understandable, as our approach neither involves specialized training for any specific attack nor requires additional model optimization. Notably, our method outperforms others in detecting CW and DF attacks that are distinctly different from FGSM attack and demonstrates stronger robustness in the worst-case scenario. This indicates that our approach has better generalizable robustness to unknown attacks.}
\begin{table*}[t]
\setlength{\tabcolsep}{3mm}
\caption{\textcolor{black}{Comparisons of our detection strategy with previous detection methods on CIFAR10 dataset.}}
\begin{center}
\begin{tabular}{|c|c|cccccccc|}
\hline
 Backbone&Method      & \multicolumn{1}{c}{FGSM} & \multicolumn{1}{c}{PGD} & \multicolumn{1}{c}{FGSM-T} & \multicolumn{1}{c}{PGD-T} & \multicolumn{1}{c}{CW} & \multicolumn{1}{c}{DF} & \multicolumn{1}{c}{AA} & \multicolumn{1}{c|}{Worst} \\ \hline
\multirow{5}{*}{ResNet18}
&LID         & 95.79                    & 86.10                   & 86.97                      & 84.71                     & 87.00                  & 76.29                  & 87.14                  & 76.29                      \\
&Mahalanobis & 99.72                    & 88.14                   & 88.28                      & 88.24                     & 82.64                  & 82.23         & 84.86                  & 82.23                      \\
&LiBRe       & 72.07                    & 91.31                   & 73.00                      & 67.70                     & 88.63                  & 57.04                  & 58.38                  & 58.38                      \\
&EPS-AD      & \textbf{100.00}          & \textbf{100.00}         & \textbf{100.00}            & \textbf{100.00}           & 78.20                  & 75.90                  & \textbf{99.23}         & 75.90                      \\
&ours        & 82.64                    & 88.42                   & 83.31                      & 88.43                     & \textbf{88.02}         & \textbf{88.44}         & 88.06                  & \textbf{82.64}             \\ \hline
\end{tabular}
\end{center}
\label{table5}
\end{table*}

\subsection{Ablation analysis}
\textbf{Analysis of Different Components in REAL.}
In defense phase, we propose a Max-Min entropy optimization scheme and an attack-aware weighting mechanism. In addition, we propose a heuristic selection strategy (HSS) of rectification rounds. We verify the effectiveness of these parts on CIFAR10. Results are shown in Table \ref{table4}, from which we observe that employing Max-Min entropy optimization (+max-min) can improve the rectification effect of samples to a certain extent. After adding the heuristic selection strategy (+HSS), the rectification accuracy can be significantly improved. In addition, we have seen that after adding attack-aware weighting mechanism (+$\beta$), rectification effect can be further improved.
\begin{figure}[t]
\centering
\includegraphics[width=\linewidth]{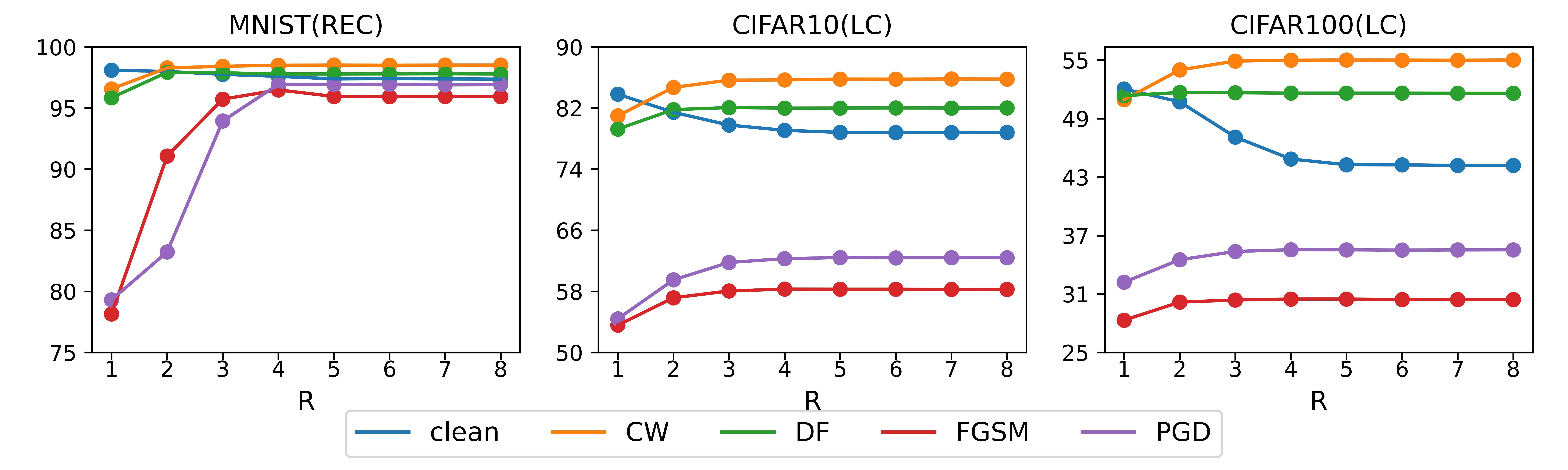}
\caption{Hyper-parameters analysis for the number  \textit{R} of rectification rounds in Algorithm \ref{algorithm}. The accuracy evaluation (\%) is conducted on MNIST, CIFAR10 and CIFAR100.}
\label{fig7}
\end{figure}

\textbf{Analysis of the hyper-parameters \textit{R} in Algorithm \ref{algorithm}.}
In our rectification algorithm, to achieve sufficient purification and correct malicious pixels of adversarial samples, an adaptive selection of number of iterative rounds was implemented. When a sample meets rectification cutoff condition in Eq. \ref{eq10} or reaches the maximum number of rectification iterations \textit{R}, the algorithm stops. And in actual experiments, we choose \textit{R}=5. To explore the ultimate impact of \textit{R} on experimental results, we have depicted ablation analysis results on three datasets. Fig. \ref{fig7} demonstrates that within the range of \textit{R} from 1 to 5, the adversarial robustness increases gradually, indicating the effectiveness of adversarial sample rectification. However, as \textit{R} continues to increase, the rectification accuracy stabilizes, suggesting that the upper limit of rectification has reached and most samples meet the correction termination criteria. Thus, we have chosen \textit{R} = 5 for all datasets in experiments.
\subsection{\textcolor{black}{Integration with diffusion purification models}}
\textcolor{black}{To further validate the effectiveness and compatibility of our method, we integrate it with the diffusion purification model. ADP\cite{yoon2021adversarial} is the first approach to utilize diffusion framework for adversarial purification. It proposes an adversarial purification method based on an Energy-Based Model (EBM) trained using Denoising Score Matching (DSM). During the testing phase, adversarial samples are purified by iteratively updating them over \textit{T} steps using the learned score function. We incorporate a two-stage entropy rectification strategy into ADP. Specifically, during the ADP purification process, entropy maximization gradients are applied during the first \textit{T/2} steps, while entropy minimization gradients are applied during the latter \textit{T/2} steps. Additionally, we incorporate an attack-aware weighting mechanism to dynamically regulate the contribution of the entropy gradient components, ultimately forming an improved hybrid strategy.
Furthermore, to comprehensively evaluate the proposed strategy, we compare it with state-of-the-art purification methods, including DiffPure\cite{nie2022diffusion}, Defense Transformer \cite{li2023learning} and RDC \cite{chenrobust}. On CIFAR10 dataset, we use ResNet18 and WideResNet28-10 as the backbone network, except for the RDC method, which is based on the U-Net diffusion classifier architecture. To ensure a fair comparison, we reproduce ADP, DiffPure and RDC under the same attack settings. As shown in Table \ref{table6}, integrating our method into ADP further improves performance, with a significant gain observed under FGSM attack. For instance, on the ResNet architecture, the performance increases by approximately 16\%. Moreover, the overall results demonstrate that incorporating our strategy into ADP significantly improves performance across various attacks, surpassing other methods.}
\begin{table*}[t]
\setlength{\tabcolsep}{3mm}
\caption{\textcolor{black}{Comparisons with other purification methods on CIFAR10 dataset.}}
\begin{center}
\begin{tabular}{|c|c|ccccccccc|}
\hline
Backbone  & Methods   & Natural   & FGSM  & PGD   & FGSM-T & PGD-T & CW & DF  & AA   & Worst     \\ \hline
Unet  & RDC   &89.85   & 73.60    &87.91   & 84.20   &86.81  &88.46     &78.00    &87.12  &73.60   
\\ \hline
\multirow{4}{*}{ResNet18}      
& Defense transformer & 87.11  & \textbf{94.38} & 91.77     & 87.55   & 84.96 & 88.65  & 88.18    & 88.57  & 84.96         
\\
& DiffPure            & 89.60          & 88.90          & 88.60          & 89.70          & 90.20          & 91.00          & 91.20          & 88.90          & 88.60          \\
 & ADP                 & \textbf{92.82} & 76.14          & 86.84          & 79.88          & 90.35          & 91.91          & 91.91          & 92.24          & 76.14          \\
 & ADP+ours            & 92.74          & 92.53          & \textbf{92.60} & \textbf{92.13} & \textbf{92.24} & \textbf{92.53} & \textbf{92.60} & \textbf{92.59} & \textbf{92.13} \\ \hline
\multirow{4}{*}{WidresNet28-10} 
& Defense transformer & 89.74          & 85.70          & 87.46          & 86.53          & 88.50          & 90.19          & 89.53          & 86.91          & 85.70          \\
& DiffPure            & 89.50          & 88.70          & 89.50          & 88.80          & 90.30          & 90.60          & 90.60          & 88.80          & 88.70          \\
 & ADP                 & \textbf{92.62} & 72.99          & 83.93          & 80.43          & 91.30          & 91.82          & 91.91          & 86.14          & 72.99          \\
 & ADP+ours            & 92.57          & \textbf{92.54} & \textbf{92.44} & \textbf{92.38} & \textbf{92.42} & \textbf{92.54} & \textbf{92.54} & \textbf{92.29} & \textbf{92.29} \\ \hline
\end{tabular}
\label{table6}
\end{center}
\end{table*}
  
\subsection{Visualization experiments}
\begin{figure}[t]
\centering
\includegraphics[width=\linewidth]{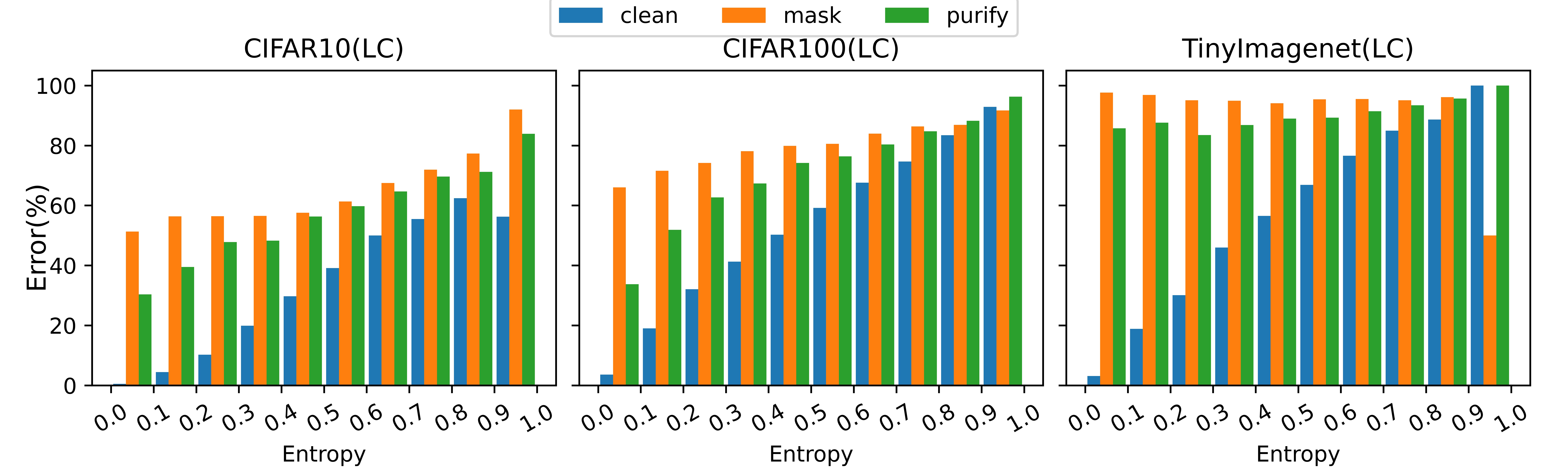}
\caption{The relationship between prediction entropy and error rate of mask sample after reverse rectification stage and purified sample after forward rectification stage. Relation consistency between clean and purified samples is better aligned.}
\label{fig8}
\end{figure}

\textbf{Analysis of relationship between prediction entropy and error rate.}
Furthermore, we delineate the statistical correlation between entropy and error rates for samples across the reverse rectification and forward rectification stages, as depicted in Fig. \ref{fig8}. Our analysis encompasses the CIFAR10/100 and TinyImageNet datasets, leveraging ResNet18 as backbone and employing LC as the auxiliary task. The adversarial perturbations were induced using PGD-20 attack which is consistent with Fig. \ref{fig1}. After the initial reverse rectification stage, the mask samples exhibit a notable decrease in error rates within the low-entropy and high-confidence regions, marking a significant departure from the adversarial sample behavior previously depicted in Fig. \ref{fig1}. Subsequently, the forward rectification in the subsequent phase serves to refine samples, aligning their entropy profiles more closely with those of clean samples. This progression underscores the efficacy of our two-stage rectification methodology.

\begin{figure}[t]
\centering
\includegraphics[width=\linewidth]{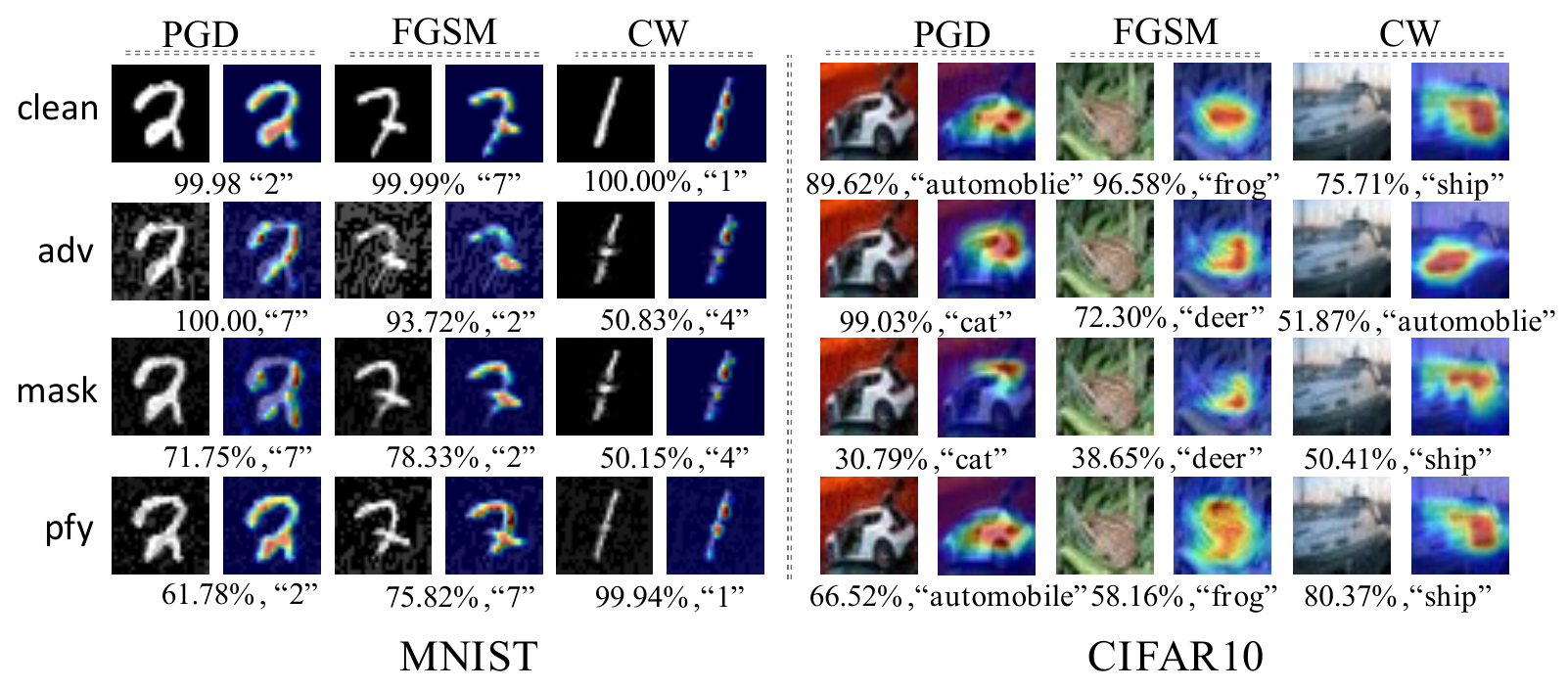}
\caption{The visual results of \textit{reverse} rectification stage (i.e., mask images) and \textit{forward} rectification stage (i.e., pfy images) on MNIST and CIFAR10 datasets. Three popular adversarial attacks, such as PGD, FGSM and CW are tested. For better visualization, the images (clean, adv, mask and pfy) together with the attention maps are shown.}
\label{fig9}
\end{figure}

\textbf{Grad-CAM Visualization.}
For a more intuitive understanding of the effects of the reverse rectification and forward rectification stages, we provide some visual results in Fig. \ref{fig9}. We sample images from MNIST dataset with CNN architecture for the REC task and from CIFAR10 dataset with ResNet18 architecture for LC task. After reverse rectification stage, it can be observed that the confidence of predictions for incorrect class gradually decreases. In the attention maps, focus on the incorrect class is disrupted, demonstrating that the mask stage plays a role in masking adversarial samples. Subsequently, after forward rectification stage, high-confidence predictions for the correct class gradually recover, and the attention maps progressively approach those of clean samples. 
\subsection{\textcolor{black}{Results of computational complexity}}
\textcolor{black}{
Since REAL is a test-time defense method, evaluating its inference time during testing is crucial. We compare the inference time across four datasets, including MNIST, CIFAR10, CIFAR100, and TinyImageNet, with the batch size set to 1 during inference. The results are shown in the Table \ref{table7}.
The results indicate that incorporating our method into SOAP increases the inference time by approximately fivefold. This is due to the five correction iterations we set, leading to a roughly fivefold increase in time complexity. As shown in Fig. \ref{fig7}, within 1 to 5 correction iterations, the correction accuracy improves significantly as the number of iterations increases. This highlights a trade-off between inference time and correction accuracy in our method. We aim to identify an optimal number of correction iterations to better balance accuracy and inference efficiency.
Notably, compared to Pixel-Defend, our method consumes less inference time. Furthermore, as shown in Table \ref{table1}, our method performs significantly better on the CIFAR10 dataset than Pixel-Defend. These results highlight the strengths of our method in providing substantial improvements in correction accuracy while maintaining a lower computational cost, making it a practical and efficient solution.}
\begin{table}[t]
\caption{\textcolor{black}{Purification time (i.e., inference
time) of the purifiers. Purification time is measured with batch size one.}}
 \begin{center}
\begin{tabular}{c|cccc}
\hline
\multicolumn{1}{c|}{\multirow{2}{*}{Method}} & \multicolumn{4}{c}{Purification Time (sec/img)}                                                \\ \cline{2-5} 
\multicolumn{1}{c|}{}                        & \multicolumn{1}{c}{MNIST} & \multicolumn{1}{c}{CIFAR10} & \multicolumn{1}{c}{CIFAR100} & \multicolumn{1}{c}{TinyImageNet} \\ \hline
Defense-GAN     &  -        & 0.13        & 0.14   & 0.31        \\
Pixel-Defend      & -     & 40.54      & 40.96  & 166.31    
 \\SOAP (LC)     & 0.06     & 0.69     & 0.69        & 0.71       \\
SOAP+ours (LC)    & 0.38  & 3.03     & 3.10      & 3.17     \\ \hline
\end{tabular}
\end{center}
\label{table7}
\end{table}

\subsection{Results on adaptive attack}
In order to further verify the reliability of our method, we discuss powerful adaptive attack methods and explore two types of adaptive attacks. 

\begin{figure}[t]
	\centering
	\includegraphics[width=\linewidth]{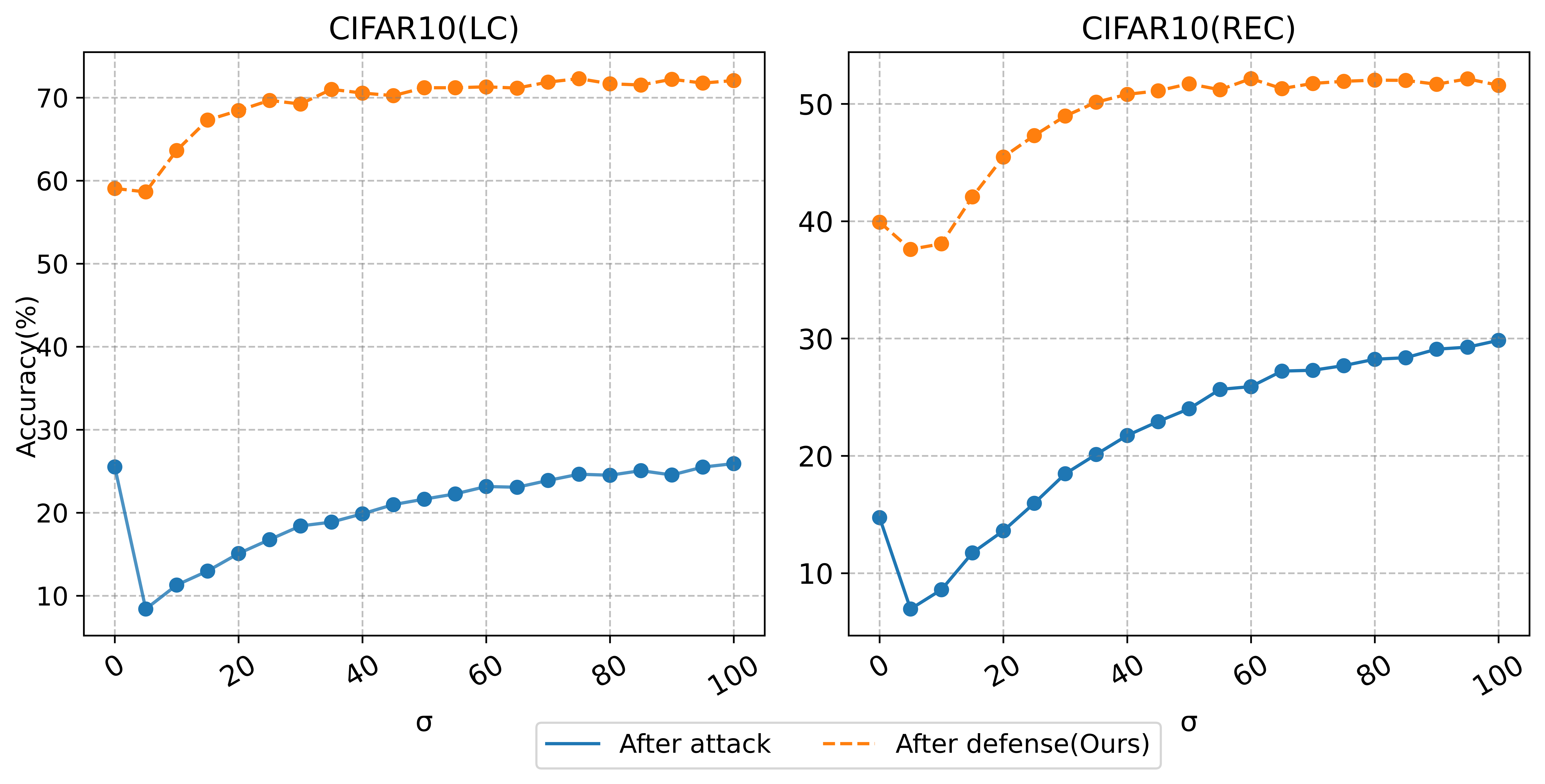}
	\caption{Results for defense-aware (adaptive) attack robustness on CIFAR10. The classification accuracy of defense-aware adversarial samples and REAL rectified samples is described. We control the intensity of the attack by changing value of $\sigma$. Two auxiliary tasks i.e., LC and REC are tested. Note that the so-called defense-aware attack means the same auxiliary task is also used in creating adversarial samples by the attacker. }
	\label{fig10}
\end{figure}

\textbf{Analysis of multi-objective attack.}
\textcolor{black}{The multi-objective attack is designed to explore whether an adversary with full knowledge of our defenses can effectively attack in a comprehensive white-box setting. To this end, we extend the standard attack loss (i.e., maximizing classification loss) by adding two defense-related components: 1) Auxiliary task loss minimization: considering that the defender may minimize auxiliary task loss to purify samples, this component is added during the attack phase to generate adversarial examples that are "on-manifold" for the auxiliary task to more effectively fool the classifier. 2) Prediction entropy maximization: countering the defense's assumption that adversarial samples exhibit low entropy. Thus, the final form of the multi-objective adaptive attack is given by Eq. \ref{eq11}, where $\sigma$ is the trade-off parameter that can be adjusted to simulate different attack strengths. } 
\begin{equation}
\textcolor{black}{	\mathcal{L}_{\mathrm{DAA}}=\mathcal{L}_{\mathrm{cls}}-\mathcal{L}_{\mathrm{aux}}+\sigma*\mathcal{L}_{\mathrm{ent}}}
	\label{eq11}
\end{equation}
\textcolor{black}{By maximizing the loss $\mathcal{L}_{\mathrm{DAA}}$, we obtain adversarial samples via adaptive attack. As shown in Fig. \ref{fig10}, the effectiveness of adaptive attack is initially improved with increasing entropy loss but decreased beyond a certain threshold. This is because in the early stages the auxiliary and entropy loss gradients align with the cross-entropy gradient, enhancing the attack power. However, when the entropy component dominates, the gradients diverge, confusing the attack and diminishing its effectiveness. This indicates that the success of adaptive attack depends on specific parameter settings, making exploitation difficult even when the defense strategy is known, highlighting the robustness of our method.
On the other hand, our method consistently demonstrates robust defense capabilities, thanks to a two-stage correction process (resembling a self-adversarial mechanism) and an attack-aware weighting mechanism. This adaptive adjustment strategy dynamically balances entropy maximization and minimization, effectively handling adversarial samples with varying attack intensities and countering high-entropy adversarial examples generated by Eq. \ref{eq11}.}

\textbf{Analysis of BPDA attack.}
Another type is the stronger BPDA attack \cite{croce2022evaluating}. \textcolor{black}{We integrate our method as a plug-and-play module into the SOAP framework. To address non-differentiable operations in SOAP's purification process, BPDA is used to approximate pipeline attacks.}
During the attack process, instead of using only gradient of model with respect to final purified image as the update direction (\textit{sign}), the average gradient of the intermediate iterations generated during the recognition process is taken. Intuitively, this directs all intermediate images towards misclassification, making the attack more effective. 
Using 1000 iterations and the parameter settings from \cite{croce2022evaluating}, we evaluate this attack on the CIFAR10 dataset.
We set the auxiliary task as label consistency (LC) and select two backbones (Resnet18 and WideResnet28-10) for this experiment. The results are shown in Table \ref{table8}. 
The results, shown in Table \ref{table8}, indicate that the original SOAP defense achieved only 3.6\% and 11.30\% accuracy under BPDA attacks. After adding our strategy, we achieve improvement effects of 15\% and 9\%, respectively. 
This suggests that even in a complete white-box attack scenario with full knowledge of our rectification process, our method still provides significant defense. 
Analyzing the underlying reasons, firstly, our defense process incorporates a Max-Min entropy optimization mechanism, making it harder for attackers to identify the gradient direction in such attacks, thus rendering the attack more difficult. Secondly, our defense method introduces an attack-aware weighting mechanism that dynamically adjusts weights with input variations, achieving real-time adaptability during test time and making it difficult to fully trace our rectification process.
\begin{table}[t]
\caption{Results for defending BPDA attack. We test the BPDA attack and defense effectiveness under label consistency auxilary task in CIFAR10 based on two backbones.}
\begin{center}
\begin{tabular}{c|cc}
			\hline
			\multirow{2}{*}{Methods} & \multicolumn{2}{c}{backbone} \\ \cline{2-3} 
			& Resnet18   & WideResnet28-10  \\ \hline
			No defense                       & 4.00       & 17.20           \\
			SOAP                             & 3.60       & 11.30           \\
			SOAP+Ours                        & \textbf{18.70}      & \textbf{20.50}           \\
			$\Delta$                           & +15.10     & +9.20           \\ \hline
\end{tabular}
\end{center}
\label{table8}
\end{table}
\subsection{Limitations}
\textbf{Auxiliary Task Choice.}
We have integrated our method into self-supervised learning-based sample rectification methods, further improving their adversarial robust generalization. Nevertheless, there still persists a trade-off issue between the accuracy of adversarial samples and clean samples (i.e., robustness vs. accuracy trade-off, a long-standing problem in adversarial defense). Hence, during the combination of our methods, to mitigate this, we devise a sample detection process integrating auxiliary loss and entropy loss. However, diverse auxiliary task designs will lead to distinct detection effects. For instance, data reconstruction task cannot be used effectively for sample detection on CIFAR10/100 datasets. Furthermore, the final rectification accuracy is also associated with the choice of auxiliary tasks. When there is a good correlation between the auxiliary tasks and the main task, it can further enhance the rectification effect. However, selecting appropriate auxiliary tasks can be challenging and requires further exploration in future work.

\begin{figure}[t]
	\centering
	\includegraphics[width=\linewidth]{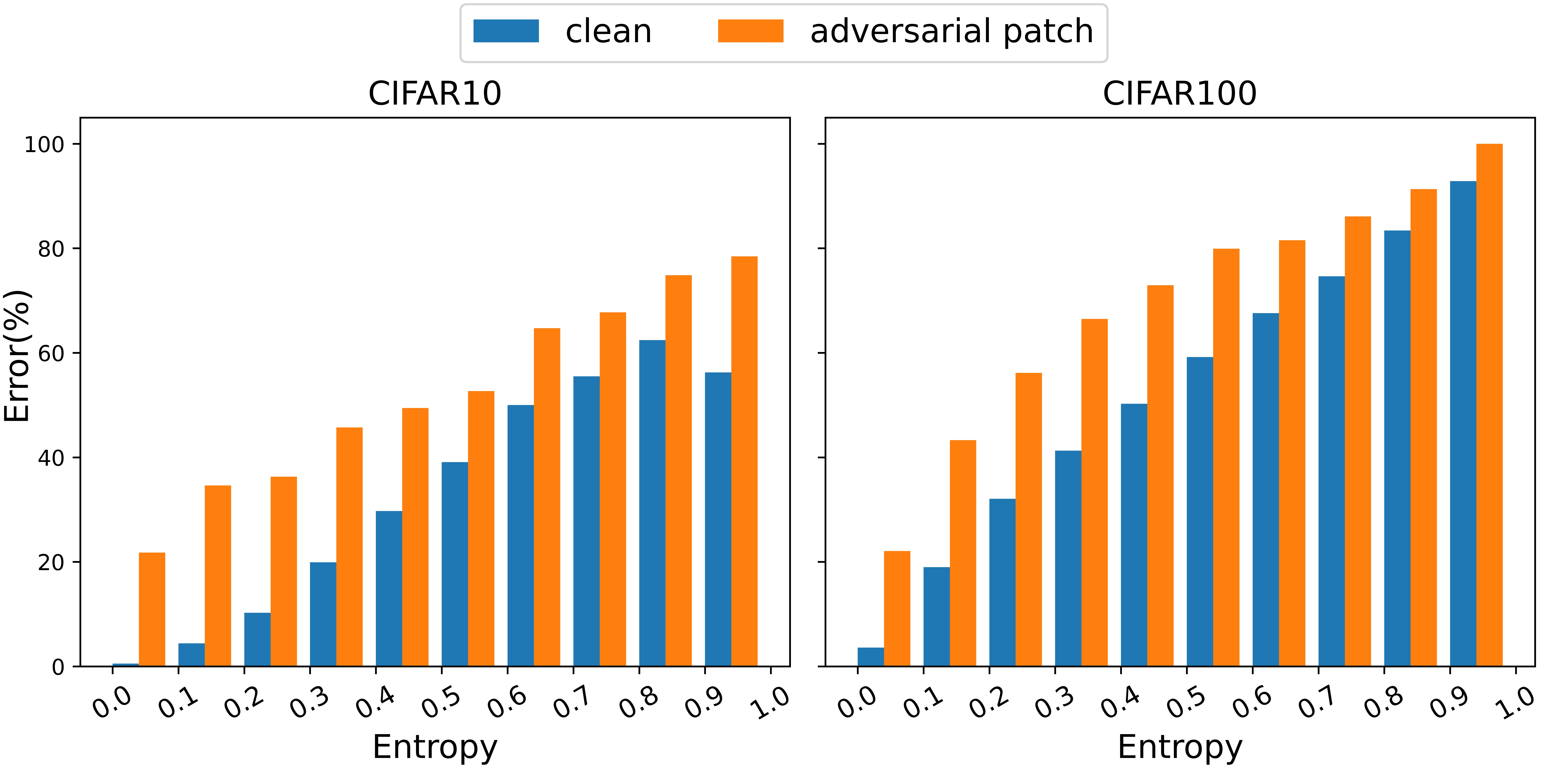}
	\caption{\textcolor{black}{The relationship between prediction entropy and error rate on physical patch attack samples.}}
	\label{fig11}
\end{figure}

\textbf{Physical Patch Attack.}
\textcolor{black}{We conduct additional experiments on the entropy properties of adversarial patches in the physical world, using CIFAR10 and CIFAR100 datasets with ResNet18 as the backbone and the patch area is set to 50\% of the original image area. As shown in Fig. \ref{fig11}, adversarial patches do not have the property of low entropy misclassification. This is rational because unlike standard examples tailored to individual images, adversarial patches are designed for generalization, enabling consistent attacks across multiple samples. Due to the need to handle diverse inputs, adversarial patches typically result in fewer low-entropy misclassifications. Additionally, we conduct defense experiments on adversarial patches of different sizes, as summarized in the Table \ref{table9}. Results indicate that existing defenses against invisible attacks are less effective against adversarial patches due to the absence of strict perturbation constraints for these patches. Improving the comprehensive robustness of existing defense models against stealth attacks and adversarial patches is an important and promising direction for future research.}
\begin{table}[t]
\caption{\textcolor{black}{Results of defending against physical patch attack. We test the physical attack and defense effects under the label consistency auxiliary task on CIFAR10 with ResNet18.}}
\begin{center}
\begin{tabular}{c|ccccc}
\hline
\multirow{2}{*}{Method} & \multicolumn{5}{c}{Attack as \% of image size} \\ \cline{2-6} 
                        & 10\%    & 20\%    & 30\%    & 40\%    & 50\%   \\ \hline
None                    & 57.00   & 43.40   & 36.50   & 21.40   & 17.60  \\
SOAP                    & 60.20   & 42.00   & 35.20   & 20.30   & 17.40  \\
SOAP+ours               & 62.00   & 46.50   & 36.80   & 20.50   & 16.20  \\ \hline
\end{tabular}
\end{center}
\label{table9}
\end{table}
\section{Conclusion and Outlook}
\label{sec:conclusion}
In this paper,  we investigate and find the low entropy prior (LE) of adversarial samples for the first time towards improving the universal generalization of adversarial robustness to unknown attacks, and further explore a reasonable Max-Min entropy optimization scheme together with an attack-aware weighting mechanism for rectifying adversarial samples. The above components can be combined with existing self-supervised sample rectification methods, resulting in a two-stage self-adversarial rectification approach, i.e. reverse rectification and forward rectification. Extensive experiments on various attacks and adaptive attacks demonstrate that our proposed method has significant effectiveness in improving the generalization of adversarial robustness.

Since this is a challenging and long-standing issue to remedy the equilibrium between ever-evolving adversarial attacks and ever-evolving defenses, in future research, further efforts should be made to explore other possible priors except LE prior among various adversarial samples for improving the generality of adversarial robustness towards open scenarios.
\bibliographystyle{plain}
\bibliography{main}
\end{document}